\documentclass[sigconf,review=false]{acmart}

\makeatletter
\def\@ACM@checkaffil{%
    \if@ACM@instpresent\else
    \ClassWarningNoLine{\@classname}{No institution present for an affiliation}%
    \fi
    \if@ACM@citypresent\else
    \ClassWarningNoLine{\@classname}{No city present for an affiliation}%
    \fi
    \if@ACM@countrypresent\else
        \ClassWarningNoLine{\@classname}{No country present for an affiliation}%
    \fi
}
\makeatother

\AtBeginDocument{%
  \providecommand\BibTeX{{%
    \normalfont B\kern-0.5em{\scshape i\kern-0.25em b}\kern-0.8em\TeX}}}

\usepackage{microtype}
\usepackage{graphicx}

\usepackage{color,soul}
\usepackage{caption,subfigure}
\usepackage{latexsym,tabularx}
\usepackage{enumitem}
\setlist{leftmargin=3mm}
\usepackage{amsfonts} %
\usepackage{tikz,amsmath,amsthm,graphicx}
\usepackage{varwidth,booktabs}
\usepackage{xcolor}
\usepackage{algorithm,algpseudocode}
\usepackage{array, makecell}
\usepackage{multirow}
\usepackage{tabularx}
\usepackage{mwe}
\usepackage{capt-of}
\usepackage{colortbl}
\usepackage{arydshln}

\newtheorem{theorem}{\textbf{Theorem}}
\newtheorem{lemma}{\textbf{Lemma}}
\newtheorem{definition}{\textbf{Definition}}
\newtheorem{proposition}{\textbf{Proposition}}
\newtheorem{property}{\textbf{Property}}

\newcommand{\m}{\textsc{FairSP}}

\setcopyright{acmcopyright}
\copyrightyear{2023}
\acmYear{2023}

\acmPrice{15.00}

\begin{document}

\title[When Fairness Meets Privacy: Fair Classification with Semi-Private Sensitive Attributes]{When Fairness Meets Privacy: \\Fair Classification with Semi-Private Sensitive Attributes}

\author{Canyu Chen}
\affiliation{
  \institution{Illinois Institute of Technology}
}
\email{cchen151@hawk.iit.edu}

\author{Yueqing Liang}
\affiliation{
  \institution{Illinois Institute of Technology}
}
\email{yliang40@hawk.iit.edu}

\author{Xiongxiao Xu}
\affiliation{
  \institution{Illinois Institute of Technology}
}
\email{xxu85@hawk.iit.edu}

\author{Shangyu Xie}
\affiliation{
  \institution{Illinois Institute of Technology}
}
\email{sxie14@hawk.iit.edu}

\author{Ashish Kundu}
\affiliation{
  \institution{Cisco Research}
}
\email{ashkundu@cisco.com}

\author{Ali Payani}
\affiliation{
  \institution{Cisco Research}
}
\email{apayani@cisco.com}

\author{Yuan Hong}
\affiliation{
  \institution{University of Connecticut}
}
\email{yuan.hong@uconn.edu}

\author{Kai Shu}
\affiliation{
  \institution{Illinois Institute of Technology}
}
\email{kshu@iit.edu}

\begin{abstract}

Machine learning models have demonstrated promising performance in many areas. However, the concerns that they can be biased against specific demographic groups hinder their adoption in high-stake applications. Thus, it is essential to ensure fairness in machine learning models. Most previous efforts require direct access to sensitive attributes for mitigating bias. Nonetheless, it is often infeasible to obtain large-scale users' sensitive attributes considering users' concerns about privacy in the data collection process. Privacy mechanisms such as local differential privacy (LDP) are widely enforced on sensitive information in the \textit{data collection stage} due to legal compliance and people's increasing awareness of privacy. 
Therefore, a critical problem is \textit{how to make fair predictions under privacy}. 
We study a novel and practical problem of \textit{fair classification in a semi-private setting}, where most of the sensitive attributes are private and only a small amount of clean ones are available.  
To this end, we propose a novel framework \textbf{{\m}} that can achieve \underline{Fair} prediction under the \underline{S}emi-\underline{P}rivate setting. First, {\m} \textit{learns to correct} the noise-protected sensitive attributes by exploiting the limited clean sensitive attributes. Then, it jointly models the corrected and clean data in an adversarial way for debiasing and prediction. Theoretical analysis shows that the proposed model can ensure fairness under mild assumptions in the semi-private setting. Extensive experimental results on real-world datasets demonstrate the effectiveness of our method for making fair predictions under privacy and maintaining high accuracy.

\end{abstract}

\maketitle

\section{Introduction}

\begin{figure}[!tbp]
    \centering
        \includegraphics[width=.47\textwidth]{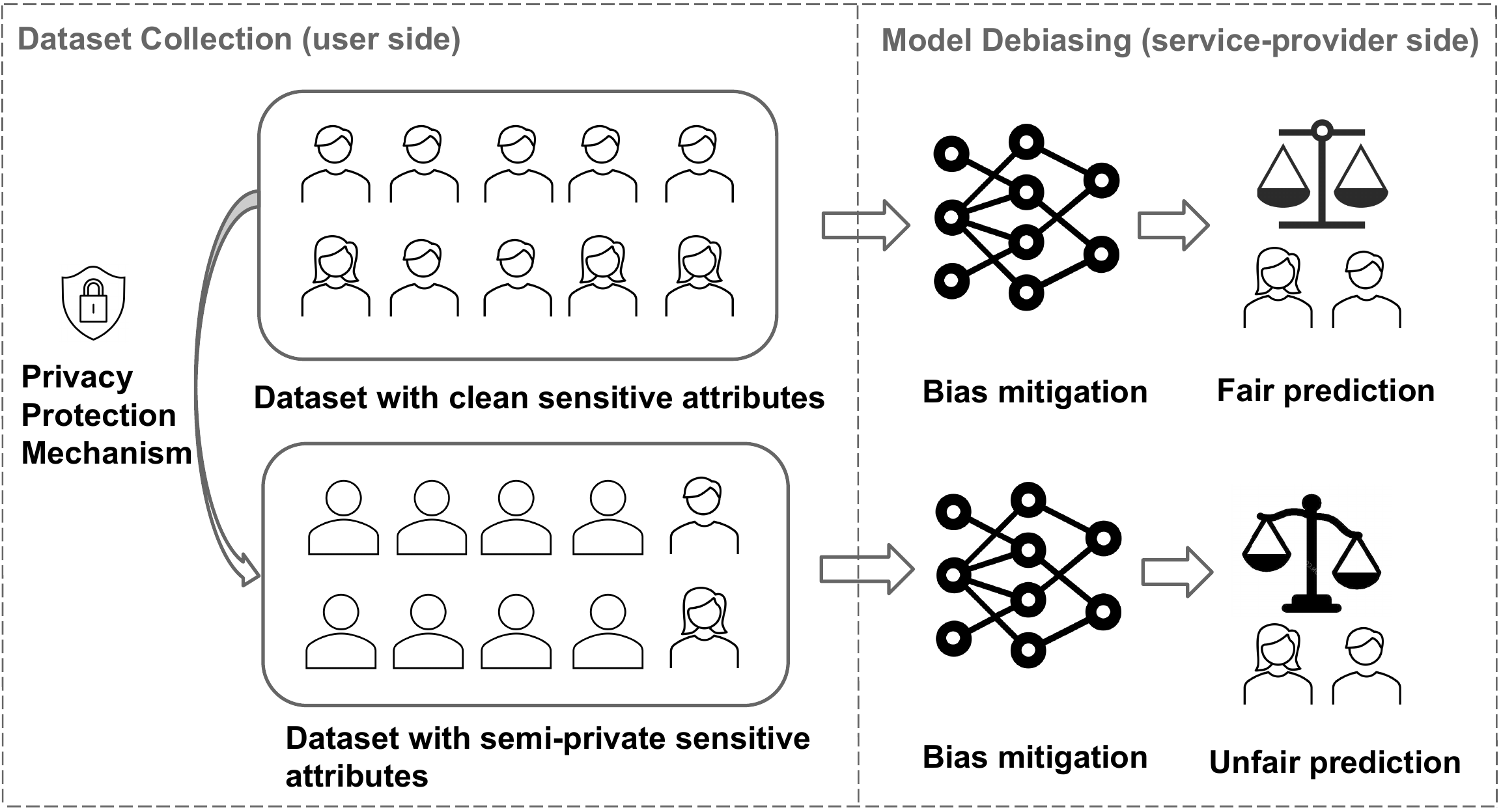}
	\vspace{-0.2cm}
    
    \caption{Comparison of the conventional setting and our proposed \textit{semi-private} setting for bias mitigation, where \textit{most sensitive attributes} are protected under certain privacy protection mechanism in the \textit{Dataset Collection} stage. Thus, bias mitigation methods that need direct access to sensitive attributes (e.g., gender) cannot achieve satisfying fairness performance with the \textit{semi-private} sensitive attributes.}
    \label{fig:semi-private}
	\vspace{-0.5cm}
\end{figure}

Machine learning has shown promising performances in various high-stake applications such as face recognition~\cite{klare2012face}, healthcare~\cite{rashid2022artificial} and loan application filtering~\cite{hamid2016developing}. However, in these applications, an emerging concern is that the prediction derived from machine learning models can often be biased and unfair to  specific (and often marginalized) groups~\cite{Hale_Kori_Bias,sun2022negative,feng2022fair}.
Such  discrimination can have detrimental societal effects that weaken the public trust among individuals, groups and the society. 
Therefore, it is critical to ensure fairness in machine learning for social good.

Recently, fair machine learning has attracted increasing attention~\cite{mehrabi2021survey,kamiran2012data}. The majority of these methods require  direct access to \textit{sensitive attributes} (e.g., race, gender, age) to preprocess the training data, regularize the model training or post-process the prediction results to derive fair predictions~\cite{madras2018learning,dwork2018decoupled,kamiran2012data}. 
However, it is often infeasible to obtain large-scale sensitive attributes for bias mitigation considering privacy.  \textbf{Privacy protection mechanisms} such as local differential privacy (LDP)~\cite{cormode2018privacy}, anonymization~\cite{10.1109/69.971193} or encryption~\cite{10.1145/3436755} are widely adopted on sensitive information in the \textit{Dataset Collection} stage due to people's increasing awareness of privacy and the legal compliance such as Electronic Communications Privacy Act (ECPA)\cite{ECPA} and General Data Protection Regulation (GDPR)\cite{GDPR}.

In practice, we observe that \textbf{it is often possible to get access to a small amount of sensitive attributes while keeping the others private}~\cite{hu2022learning,li2022occupant,DBLP:journals/popets/HilsWB21,bogen2020awareness,9162015, klemperer1989supply}. From \textit{user} perspective, previous research has shown that the majority of people can accept that researchers and relevant parties approved by IRB (Institutional Review Board) have access to their sensitive data~\cite{li2022occupant}. From \textit{company} perspective, there is also previous work that illustrates private companies can collect or infer sensitive attribute data to pursue antidiscrimination goals under the U.S. civil rights law in the domains of credit, employment, and healthcare~\cite{bogen2020awareness}.

Therefore, the following research question arises: \textbf{\textit{Can we achieve fair prediction with mostly private sensitive attributes?}} As shown in Figure~\ref{fig:semi-private}, we study a practical and novel problem of \textbf{\textit{fair classification in a semi-private setting}}, where most of the sensitive attributes (e.g., gender) are private and only very limited clean ones are available when datasets are collected from users. In this paper, we consider the scenario where the datasets have most of the sensitive attributes protected under the \textbf{Local Differential Privacy (LDP)} mechanism since it is widely adopted in various applications~\cite{bindschaedler2017plausible,Qu2021NaturalLU,kasiviswanathan2008learn,cormode2018privacy,carvalho2022survey,xiong2020comprehensive}, provides strong privacy guarantee without the assumption of a trusted third-party service provider (see Section~\ref{sec:Impact_of_Privacy_on_Fairness}) and has advantageous properties (see Section~\ref{sec:Discussions_on_Privacy_Guarantee}). We will further explore other privacy mechanisms in the future.

However, it is nontrivial to build fair machine learning models with mostly private sensitive attributes. \textit{First}, private sensitive attributes are noisy, and directly applying conventional debiasing techniques on them can lead to sub-optimal performances. Some initial efforts have verified that noise-protected sensitive attributes may hurt the performance of conventional debiasing models~\cite{lamy2019noise,wang2020robust}.
\textit{Second}, it is unknown whether or not it is helpful to incorporate the limited clean sensitive attributes. Such limited instances with clean sensitive attributes are inadequate for training a fair classification model directly because the model can easily overfit the small dataset. 
\textit{Third}, the conventional model design cannot effectively leverage both the mostly private sensitive attributes and limited clean ones. Most previous debiasing models would treat each instance the same way when applied to our proposed semi-private setting~\cite{agarwal2018reductions,bechavod2017penalizing,mehrabi2021survey}. Since the clean sensitive attributes are very limited, it is important to explore how to effectively exploit them to enhance fairness in the semi-private setting.

Therefore, to address these challenges, we first conduct a pilot study on \textbf{the impact of privacy on fairness} under the local differential privacy (LDP) definition. We found that the fairness performance of debiasing models is enhanced with a smaller noise rate on the sensitive attributes, which is determined by a constant \textit{privacy budget}. 
Motivated by the finding, we propose a novel framework \textbf{{\m}} for fair classification with semi-private sensitive attributes, which \textbf{\textit{learns to correct}} the noisy sensitive attributes  by exploiting the limited clean ones. Specifically, {\m} first estimates the \textit{Sensitive Attribute Corruption Matrix} for the noisy sensitive attributes by leveraging both  the noisy ones and very limited clean ones. Then, it learns a \textit{Private Sensitive Attribute Corrector} with the matrix to estimate the ground-truth distribution of the noisy sensitive attributes. 
Finally, {\m} conducts \textit{Semi-Private Adversarial Debiasing} with both the corrected and clean data. 
It is worth noting that {\m} does not impact the privacy guarantee provided by LDP in the dataset collection stage due to its key property of \textbf{\textit{immunity to post-processing}} (see Section~\ref{sec:Discussions_on_Privacy_Guarantee} for more details).

In summary, our main contributions are as follows:
\begin{itemize}
\item We study a novel and practical problem of fair classification with semi-private sensitive attributes.
\item We empirically study the impact of privacy on fairness under the definition of local differential privacy (LDP) mechanism.
\item We propose a new end-to-end framework {\m} which simultaneously derives corrected sensitive attributes from private ones and learns a fair classifier in an adversarial way with both the clean and corrected data.  
\item We conduct a theoretical analysis on when fairness can be achieved in our proposed semi-private setting.
\item We perform extensive experiments on real-world datasets to validate the effectiveness of the proposed
\textit{learning to correct} method for fair classification in the semi-private setting.
\end{itemize}

\begin{figure*}[htbp!]
	\centering
	\subfigure[ADULT]{
	\includegraphics[width=0.24\textwidth]{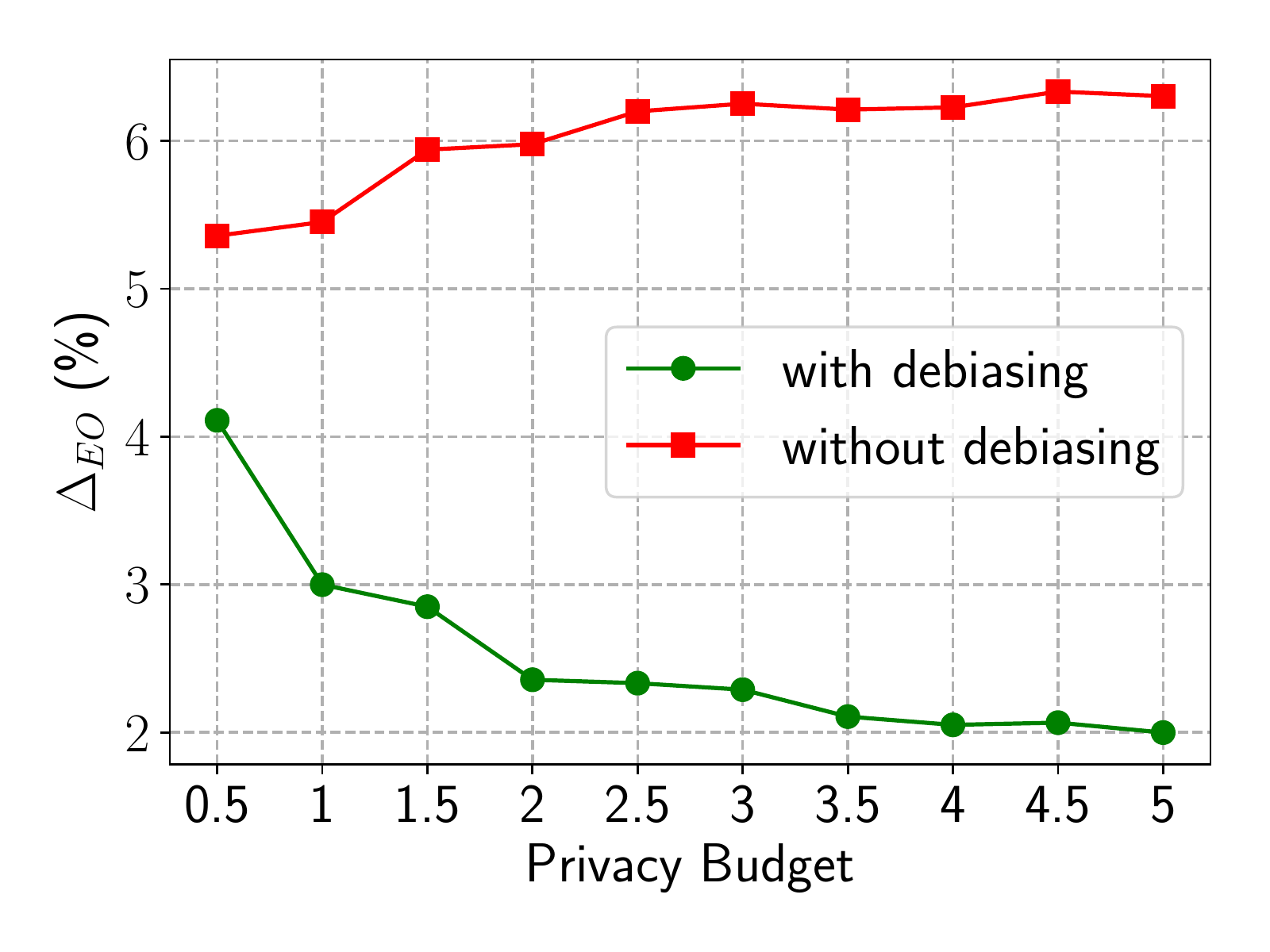}}
		\subfigure[ADULT]{
	\includegraphics[width=0.24\textwidth]{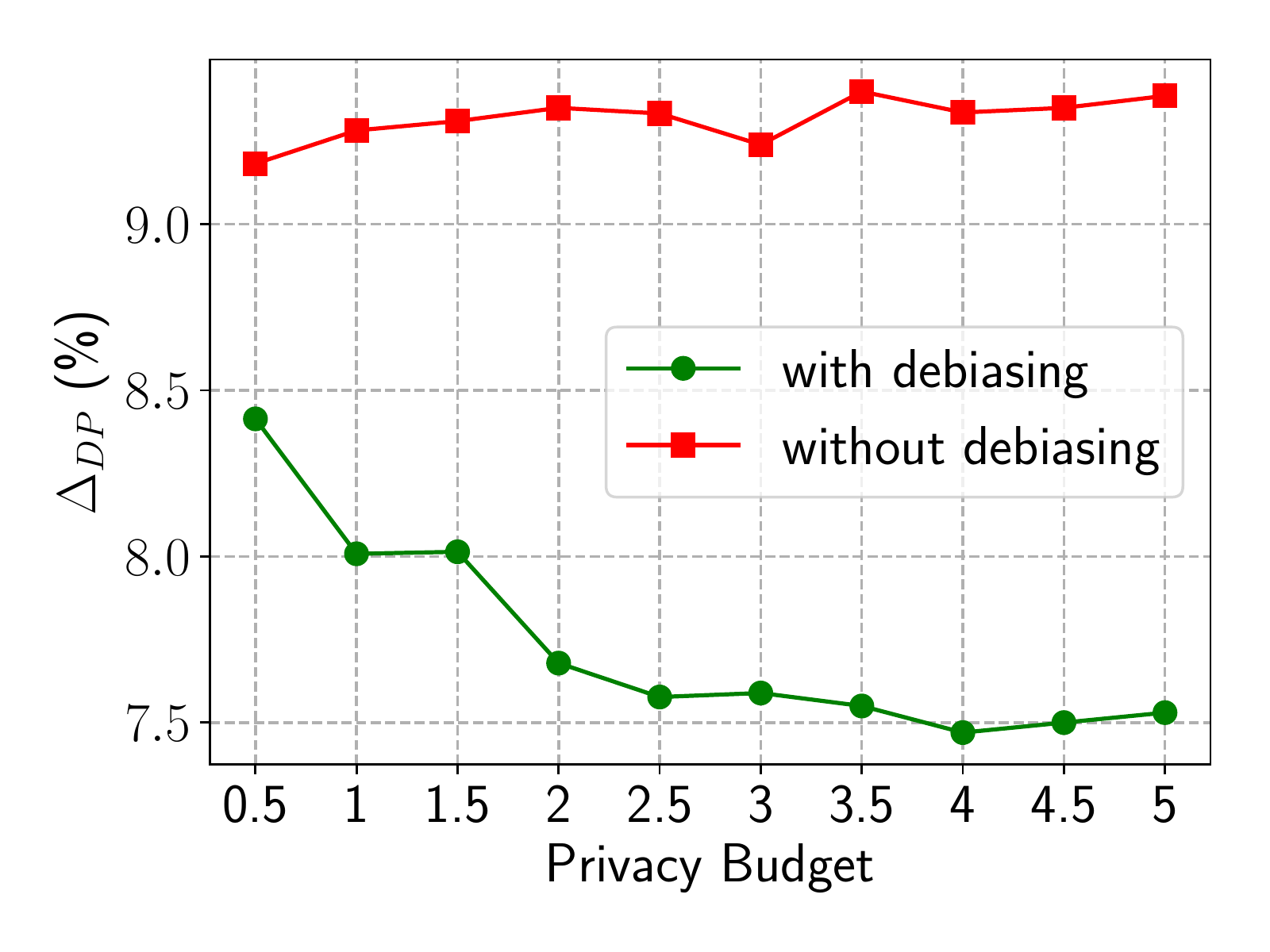}}
		\subfigure[COMPAS]{
	\includegraphics[width=0.24\textwidth]{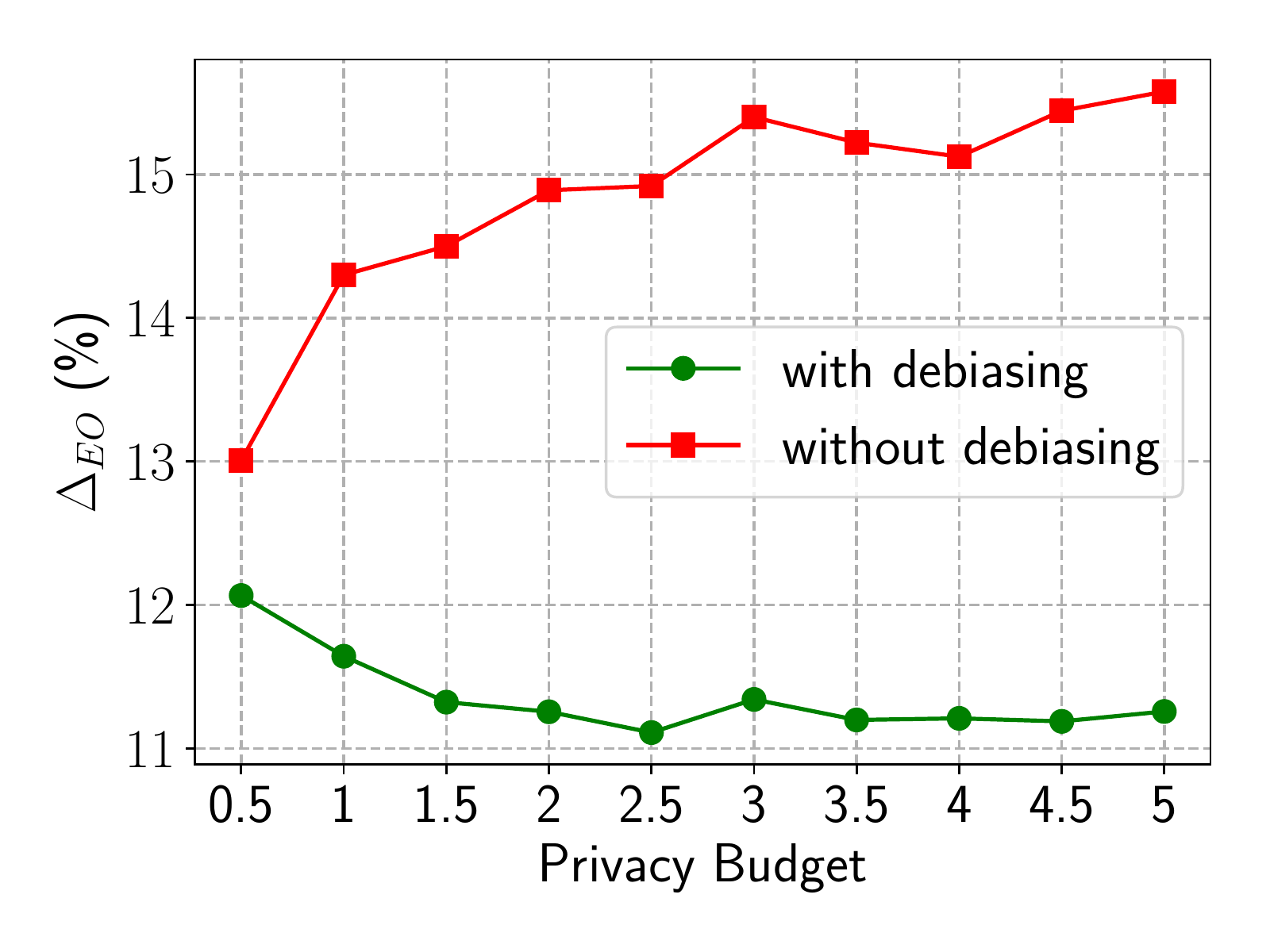}}
			\subfigure[COMPAS]{
	\includegraphics[width=0.24\textwidth]{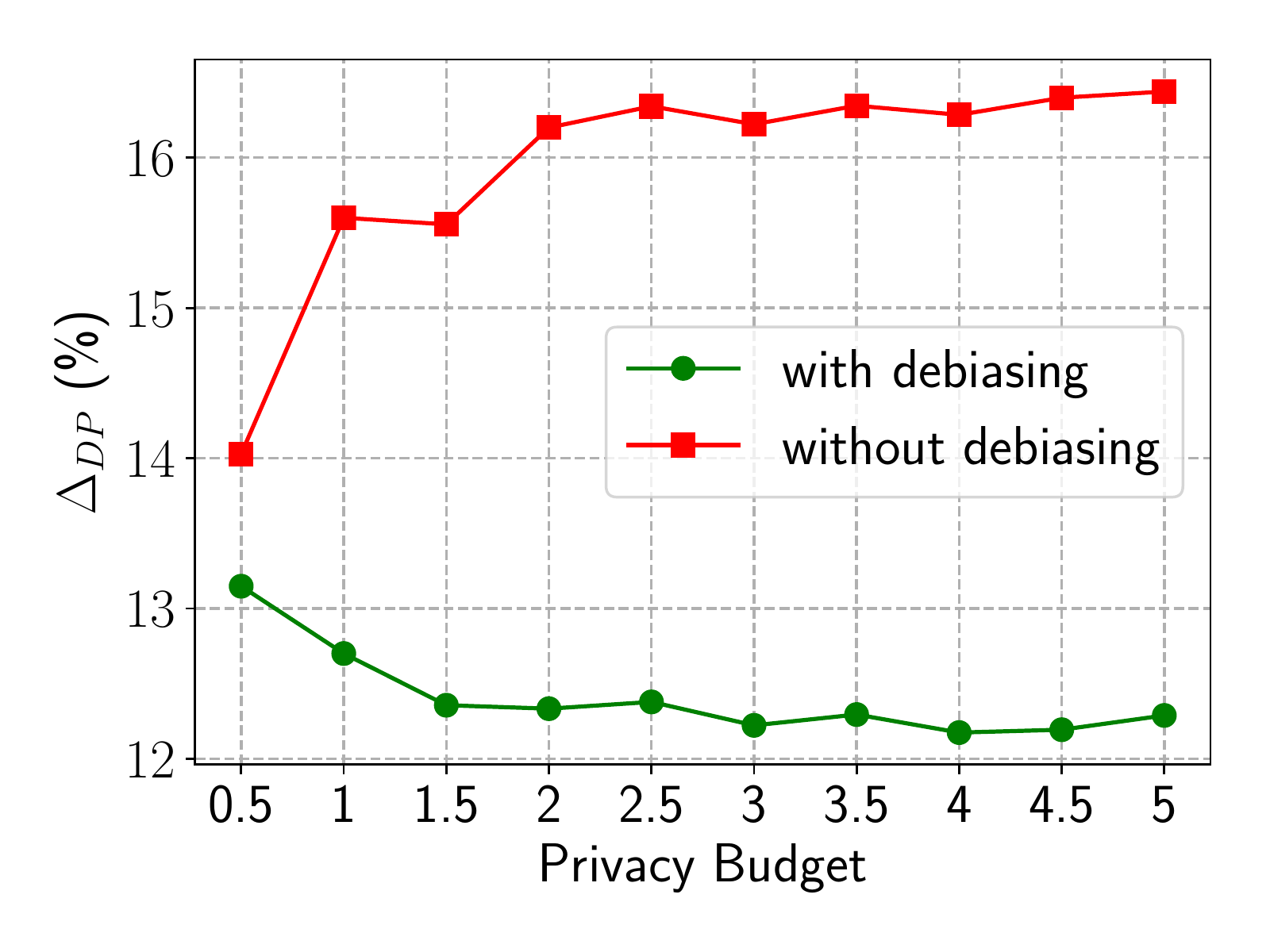}}
 	\vspace{-0.4cm}
	 \caption{Assessing the impact of privacy on fairness under \textit{Local Differential Privacy} (LDP) definition. Results of ``with debiasing'' ({\sf green line})   show \textit{stronger} privacy guarantee (privay budget $\downarrow$) leads to \textit{worse} debiasing performance ($\Delta_{EO}$ and $\Delta_{DP}$ $\uparrow$).
}
	\label{fig:prelimi}
\end{figure*}

\section{Assessing the Impact of Privacy on Fairness}
\label{sec:Impact_of_Privacy_on_Fairness}

In this section, we first briefly introduce the background of privacy protection mechanisms and LDP. Then we conduct a preliminary  study to assess the impact of LDP on fairness performances.

\subsection{Privacy Protection Mechanism}
\label{section:LDP}
Differential privacy (DP) is a typical privacy protection mechanism in prior works~\cite{garrido2022lessons}, which provide a guarantee that the query results are \textit{indistinguishable} for two datasets that only differ in one entry~\cite{10.1561/0400000042}. DP is originally adopted for statistical datasets~\cite{dwork2006calibrating} and also more recently widely applied for machine learning applications~\cite{gong2020survey}. However, many previous works focus on the \textit{Centralized Privacy} setting, which needs a strong assumption that there is a trusted service provider to process users' data. This is often unrealistic considering many users do not trust third-party service providers. Thus, in this paper, we consider the \textit{Local Privacy} setting, which ensures privacy protection in the data collection stage and does not need the assumption of trusted service providers. Local differential privacy (LDP)~\cite{cormode2018privacy} is a typical mechanism in this setting.

\subsection{Local Differential Privacy Mechanism}
\label{section:LDP}

The local differential privacy (LDP) mechanism provides strong guarantees in \textit{indistinguishability} aspect by directly injecting noise into each individual data entry in the dataset collection stage. The formal Local differential privacy definition is as follows:

\begin{definition}\label{definition1}
Given $\epsilon$ > 0, a randomized mechanism $\mathcal{M}: \mathcal{X} \rightarrow \mathcal{Y}$ satisfies $\epsilon$-local differential privacy, if for all possible pairs of users' private data $x_i$ and $x_j$, the following equation holds:

\vspace{-0.2cm}
 \begin{align}
\forall y \in \mathcal{Y}: \frac{P\left(\mathcal{M}\left(x_{i}\right)=y\right)}{P\left(\mathcal{M}\left(x_{j}\right)=y\right)} \leq e^{\epsilon}
 \end{align}
where Range($\mathcal{M}$) denotes every possible output of $\mathcal{M}$.
\end{definition}

The parameter $\epsilon$ denotes the privacy budget to balance the utility and privacy guarantee of the model. A smaller $\epsilon$ represents stronger privacy guarantee and weaker utility. Following the existing work~\cite{lamy2019noise}, we only apply LDP on the sensitive attributes of datasets and can obtain the following lemma:

\begin{lemma}\label{lemma1}

To achieve $\epsilon$-local differential privacy on the binary sensitive attribute, we can randomly flip the sensitive attributes with a probability of $p = \frac{1}{exp(\epsilon)+1}$.

\end{lemma}
Specifically, considering an instance $\{x_i, y_i, a_i\}$ in the dataset where $x_i$, $y_i$, $a_i$ denote non-sensitive attributes, label, and sensitive attribute of respectively and only $x_i$ disclosed by an attacker, $\hat{a}_i$ denotes the sensitive attribute of the instance after adding noise (i.e., flipping the sensitive attribute with a probability $p$). Assuming the sensitive attribute $a_i$ is queried, the attacker intends to know whether $a_i = 0$ or $a_i = 1$. To comply with the definition of LDP, the probability is calculated as follows: $p = \frac{1}{exp(\epsilon)+1}$.

The detailed proof of the Lemma refers to the Lemma 3 in  ~\cite{lamy2019noise}. Based on the Lemma, we can obtain \textit{differentially private sensitive attributes} by flipping at a certain probability. If the flipping probability satisfies the condition that $p = \frac{1}{exp(\epsilon)+1}$, then the sensitive attributes have the \textit{$\epsilon$-local differential privacy guarantee}.

\subsection{Discussions on Fairness under Privacy}
\label{sec:Discussions_on_Fairness_under_Privacy}

In this subsection, we investigate the impact of privacy on fairness under the LDP definition. 
As for the impact of LDP on fairness, we conduct two groups of preliminary experiments. For the first group of experiments, we study the impact of LDP on non-debiasing models. We take a vanilla multi-layer perceptron (MLP) network as the example. For the second group of experiments, we study the impact of LDP on debiasing models. We adopt adversarial debiasing as an example, which is a representative fairness method. We conduct each group of experiments on  datasets ADULT and COMPAS, which are two typical fairness datasets. We set six privacy budgets for each group of experiments as 0.5, 1, 1.5, 2, 2.5, 3, 3.5, 4, 4.5, 5. We follow Lemma \ref{lemma1} to implement the LDP mechanism. We utilize $\Delta_{EO}$ and $\Delta_{DP}$ as the fairness evaluation metrics
(See Section~\ref{sec:Evaluation_Metrics} for the definition). 
From Figure~\ref{fig:prelimi}, we have the following observations:

\begin{itemize}
    \item  \textbf{For vanilla models without debiasing, stronger privacy guarantee improves the fairness performance.} We can observe that with a lower privacy budget, the vanilla models have better fairness performance on $\Delta_{EO}$ and $\Delta_{DP}$. 
    With a stronger privacy guarantee, the privacy budget decreases and the flipping rate increases, which means there is more noise  injected into the sensitive attributes of the dataset.
    Thus, the vanilla models cannot learn the \textit{explicit bias} contained in the sensitive attributes. 
    \item \textbf{For debiasing models, a stronger privacy guarantee leads to worse fairness performance.}
    From Figure~\ref{fig:prelimi}, we can see that with a lower privacy budget, the debiasing models have a worse fairness performance on $\Delta_{EO}$ and $\Delta_{DP}$ for the two datasets. This is because the debiasing models need to explicitly leverage sensitive attributes for mitigating bias.  With a stronger privacy guarantee, there is a lower privacy budget and more noisy sensitive attributes, which causes the debiasing models ineffective in mitigating the \textit{implicit bias} contained in non-sensitive attributes.
\end{itemize}

\begin{figure*}[!tbp]
    \centering

    \includegraphics[width=1\textwidth]{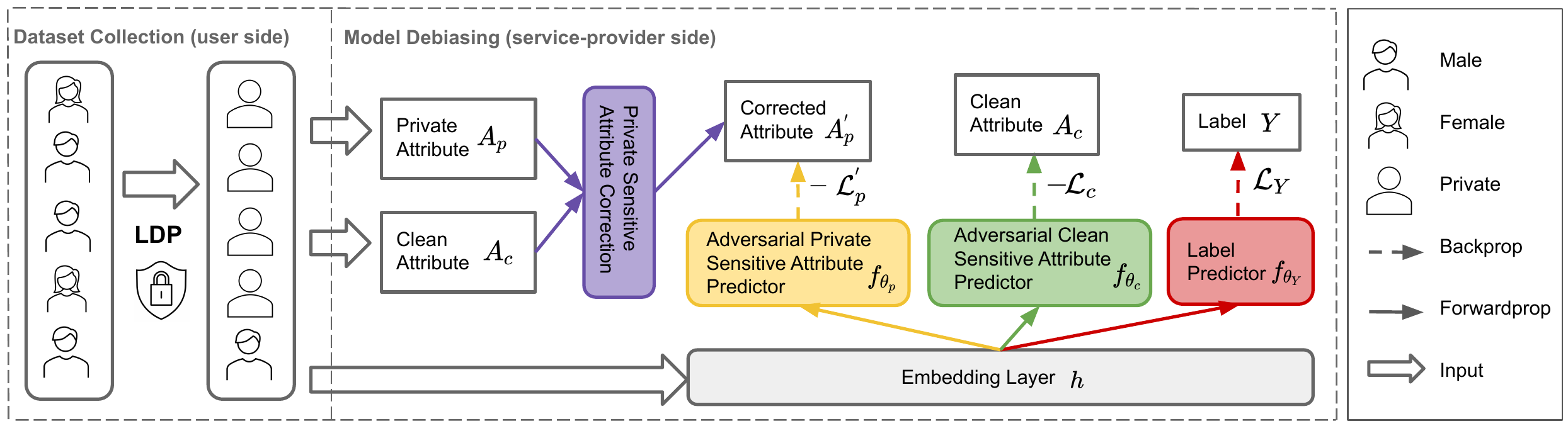}
    
	\vspace{-0.2cm}
    \caption{An illustration of the proposed framework {\m}  under the semi-private setting. In the \textit{Dataset Collection} stage, local differential privacy (LDP) mechanism is enforced on most of the sensitive attributes in the dataset. In the \textit{Model Debiasing} stage, our proposed {\m} consists of two major modules: (1) a private sensitive attribute correction module for correcting noisy sensitive attributes; and (2) a semi-private adversarial debiasing module for learning a fair classifier.
    }

    \label{fig:framework}
	\vspace{-0.3cm}
\end{figure*}

\section{Problem Statement}
\label{sec:Problem_Statement}
In this section, we give the formal problem definition of \textbf{\textit{fair classification with semi-private sensitive attributes}}. Let $\mathcal{D}=\{\mathcal{X},\mathcal{A}, \mathcal{Y}\}$ denote the dataset, where $\mathcal{X}$, $\mathcal{A}$, and $\mathcal{Y}$ represent the set of data samples excluding sensitive attributes, sensitive attributes, and labels respectively. For the sensitive attributes $\mathcal{A}$, it consists of a small number of clean ones $\mathcal{A}_c$, and a large number of private ones $\mathcal{A}_p$, i.e., $\mathcal{A}=\mathcal{A}_c\cup\mathcal{A}_p$. Usually, the number of clean sensitive attributes is much smaller than that of private sensitive attributes.

Following the existing works on fair classification~\cite{du2021fairness,zhao2021fair,lamy2019noise}, we adopt the setting with binary class and binary sensitive attribute, i.e., $A$ and $Y$ can be either $0$ or $1$. We evaluate the performance of fairness using metrics Equal Opportunity and Demographic Parity~\cite{barocas2017fairness,mehrabi2021survey} (See details in Section~\ref{sec:Evaluation_Metrics}).
The problem of fair classification with semi-private sensitive attributes is formally defined as follows:

\begin{center}
\fbox{\parbox[c]{.95\linewidth}{\textbf{Problem Statement:}
Given the training data $\mathcal{D}$ with a  limited number of clean sensitive attributes and a large number of private sensitive attributes, learn an effective classifier that generalizes well to unseen instances, while satisfying the fairness criteria such as demographic parity and equal opportunity.
}}
\end{center}

\section{Dataset Collection with Semi-Private Sensitive Attributes}
Most previous bias mitigation models need to get direct access to sensitive attributes~\cite{agarwal2018reductions,bechavod2017penalizing,mehrabi2021survey,DBLP:journals/corr/abs-2010-04053,zhang2018mitigating}, but it is increasingly difficult to obtain large-scale datasets with sensitive information for training models considering users' concerns about privacy in the \textit{data collection stage}. Therefore, we study a more practical problem where most of the sensitive attributes are private and only very limited ones are clean since it is often possible to obtain a small amount of clean sensitive attributes
~\cite{hu2022learning,li2022occupant,DBLP:journals/popets/HilsWB21,bogen2020awareness,9162015, klemperer1989supply}. 
To better reflect the reality of datasets collected from users, we assume only 20\% to be non-private and 80\% to be enforced with the local differential privacy mechanism among all the samples in the setup of our main experiments (see Section~\ref{sec:Performance_of_Fair_Classification}). Furthermore, we also demonstrate the effectiveness of our method when the non-private percentage of the dataset is extremely low  such as 2\%, 0.2\% and 0.02\% (see Section~\ref{sec:Private_Data_Ratio}). 
The  non-private samples provide clean sensitive attributes and will not overlap with the private ones.

\section{{\m}: Fair Classification with Semi-private Sensitive Attributes}
\label{section_model}
Under the semi-private problem setting, we propose a novel framework {\m}, which effectively leverages the very limited clean sensitive attributes and mostly noisy ones for bias mitigation. We first introduce the motivation and then present the details.

\subsection{Motivation of {\m}}

Based on pilot study discussed in Section~\ref{sec:Discussions_on_Fairness_under_Privacy}, we can observe that the fairness performance is improved when the privacy noise injected into sensitive attributes decreases, which inspires us that \textbf{we can attempt to improve the fairness performance of debiasing models under the semi-private setting by reducing the noise in sensitive attributes }. 
Considering there exists a correlation between non-sensitive attributes and sensitive attributes, it is feasible to estimate the ground-truth distribution of the noisy sensitive attributes by exploiting the very limited available clean sensitive attributes and non-sensitive attributes. Then, the fairness performance of the debiasing models with the ``corrected'' sensitive attributes can be improved since their noise is likely to be decreased.

\subsection{Semi-Private Adversarial Debiasing}
In our semi-private scenario, we have two distinct types of sensitive attributes: clean and private. Our objective is to build a framework that leverages the information from both clean and private samples and learns an underlying common representation that can induce fair classification. 
As shown in Figure~\ref{fig:framework}, we propose a \textbf{Semi-Private Adversarial Debiasing} framework for bias mitigation in our setting, which consists of Embedding Layer $h(\cdot)$, Label Predictor $f_{\theta_Y}$, Adversarial Private Sensitive Attribute Predictor $f_{\theta_p}$ and Adversarial Clean Sensitive Attribute Predictor $f_{\theta_c}$.

Specifically,  we first utilize a shared encoder $h(\cdot)$  to learn the embedding vector, which is then fed into separate layers for debiasing and predicting. For Label Predictor $f_{\theta_Y}$, it aims to minimize the prediction error of labels with the following objective function:
\begin{equation}
    \min_{\theta_{h}, \theta_{Y}}  \mathcal{L}_{Y} = \mathbb{E}_{(X,Y)\in \mathcal{D}}\ell(Y, f_{\theta_Y}(h(X)))
 \vspace{-0.2cm}
\end{equation}
where $\ell$ refers to cross entropy loss, $\theta_h$ and $\theta_Y$ are the parameters for the embedding layer $h(\cdot)$ and label predictor $f_{\theta_Y}$ respectively. 

In addition, to learn fair representations and make fair predictions, we incorporate two adversaries $f_c$ and $f_p$ to predict the clean and private sensitive attributes respectively. The encoder $h(\cdot)$ tries to learn the representation that can fool the adversaries.  Thus, $f_{\theta_c}$ and $f_{\theta_p}$ are jointly optimized with the following objective function:
\begin{equation}\label{eqn:a}
\min_{\theta_{h}}\max_{\theta_c, \theta_p} \mathcal{L}_a = \mathcal{L}_c+\alpha \mathcal{L}_p
 \vspace{-0.15cm}
\end{equation}
where  $\theta_c$ and $\theta_p$ are the parameters for the adversaries predicting the clean and private sensitive attributes respectively, $\alpha$ is a hyper-parameter that controls the relative importance of the loss functions computed over the data with clean and private sensitive attributes, and $\mathcal{L}_c$ and $\mathcal{L}_p$ are defined as follows:
 \begin{align}
\min_{\theta_{h}}\max_{\theta_c} \mathcal{L}_c &= \mathbb{E}_{X \sim p(X \mid A_c=1)}[\log (f_{\theta_c}(h(X)))] \nonumber\\
&+\mathbb{E}_{X\sim p(X \mid A_c=0)}[\log (1-f_{\theta_c}(h(X)))] 
 \end{align}
  \vspace{-0.5cm}
  \begin{align}\label{eqn:p}
 \min_{\theta_{h}}\max_{\theta_p} \mathcal{L}_p &  = 
 \mathbb{E}_{X \sim p(X \mid A_p=1)}[\log (f_{\theta_p}(h(X)))] \nonumber \\
&  +\mathbb{E}_{X\sim p(X \mid A_p=0)}[\log (1-f_{\theta_p}(h(X)))] 
 \end{align}
Finally, the overall objective function of adversarial debiasing for fair classification is a minimax function,  where $\beta$ controls the importance of the adversarial sensitive attribute predictors:
 \begin{equation}\label{eqn:l2}
\min_{\theta_{h},\theta_Y}\max_{\theta_c,\theta_p} \mathcal{L}_{adv} = \mathcal{L}_Y-\beta (\mathcal{L}_c+\alpha \mathcal{L}_p)
\end{equation}

\subsection{Private Sensitive Attribute Correction}
Since private sensitive attributes are noisy, directly applying adversarial debiasing on them may lead to sub-optimal results, which is demonstrated by our preliminary experiments in Section~\ref{sec:Impact_of_Privacy_on_Fairness}. Also, the finding that reducing the noise can be beneficial for bias mitigation inspires us to consider \textbf{\textit{learning to correct}} these noisy sensitive attributes before feeding them into the adversarial predictors. Specifically, the \textbf{Private Sensitive Attribute Correction} method consists of two steps: Sensitive Attribute Corruption Matrix Estimation and Learning Private Sensitive Attribute Corrector. First, we illustrate the Conditional Independence Assumption needed in our method and then present the details.

\subsubsection{Conditional Independence Assumption}
\label{sec:Conditional_Independence_Assumption}
Let $X$ denote the distribution of samples excluding sensitive attributes (in other words, the distribution of non-sensitive attributes), the conditional independence assumption refers that the private sensitive attribute $A_p$ and non-private sensitive attribute $A_c$ are conditionally independent given $X$. This assumption generally holds in our semi-private setting with the following propositions. 

\begin{proposition}\label{pro0}
When $\epsilon$-local differential privacy is enforced on the binary sensitive attribute, the privacy noise complies with the class-conditional distribution.
\end{proposition}
\begin{proof}
From Lemma~\ref{lemma1}, each sensitive attribute $A = i$ can be flipped to a noisy sensitive attribute $A = j$ with a constant probability $p(A = j | A = i) = \frac{1}{exp(\epsilon)+1}$, where $i$ and $j$ can be 0 or 1 for binary sensitive attribute. If we denote the probability as $p$, then the noise transition matrix is $\begin{pmatrix} 
1-p & p \\
p & 1-p 
\end{pmatrix}$. Thus, the noise distribution is independent from the input and complies with the class-conditional distribution~\cite{10.1023/A:1022873112823}.
\end{proof}
The Proposition~\ref{pro0} suggests the connection between \textit{local differential privacy noise} and \textit{class-conditional label noise}. Based on this proposition, we can infer the following proposition:
\begin{proposition}\label{pro2}
When $\epsilon$-local differential privacy is enforced on the binary sensitive attributes of the dataset $\mathcal{A}$ and each sample $X$ corresponds to a single ground-truth sensitive attribute, the private sensitive attribute $A_p$ and non-private sensitive attribute $A_c$ are conditionally independent given $X$.
\end{proposition}
\begin{proof}
From Proposition~\ref{pro0}, we can see that local differential privacy noise complies with the class-conditional distribution. Also, each sample $X$ corresponds to a single ground-truth sensitive attribute. Then we can infer that $A_c$ and $A_p$ are conditionally independent given $X$ based on the proof in the Appendix A of~\cite{hendrycks2018using}.
\end{proof}
Since it is generally true that $X$ corresponds to a single ground-truth sensitive attribute in our setting, then Proposition~\ref{pro2} suggests that the conditional independence assumption generally holds.

\subsubsection{Sensitive Attribute Corruption Matrix Estimation}

Specifically, given the dataset $\mathcal{D}_c=\{\mathcal{X}_c,\mathcal{A}_c, \mathcal{Y}_c\}$ with instances containing clean sensitive attributes of $l$ categories, and $\mathcal{D}_p=\{\mathcal{X}_p,\mathcal{A}_p,\mathcal{Y}_p\}$ with instances' sensitive attributes being private. From Proposition~\ref{pro0}, we can see the connection between LDP noise and class-conditional label noise. We aim to estimate the \textbf{Sensitive Attribute Corruption Matrix} $\mathbf{C}\in \mathbb{R}^{l\times l}$ to model the sensitive attribute corruption process.  Let $\mathbf{C}_{mr}$ denote the corruption probability from  $A_c=m$ to $A_p=r$, $\hat{\mathbf{C}}$, $\hat{\mathbf{C}}_{mr}$ be the estimation of $\mathbf{C}$, $\mathbf{C}_{mr}$ respectively.

We first train a sensitive attribute predictor $g(\cdot)$ on the private data $\mathcal{D}_p$, then $g(X)$ can be an estimation of $p(A_p|X)$, i.e., $g(X)= \hat{p}(A_p|X)$. Then we can utilize $g$ and $\mathcal{D}_c$ to estimate $\mathbf{C}$.
Let $\mathcal{X}_m$ be the subset of $\mathcal{X}_c$ with sensitive attribute $A_c=m$. 
Based on the \textit{conditional independence assumption} in Section~\ref{sec:Conditional_Independence_Assumption}, we can infer that $A_p$ is conditional independent from
$A_c$ given ${X}$, i.e., $p(A_p|A_c,X) = p(A_p|X)$. Thus we can obtain:
    \begin{align}
     \mathbf{C}_{mr} = p(A_p=r|A_c=m) &\approx \frac{1} {|\mathcal{X}_m|}\sum_{X\in\mathcal{X}_m}\hat{p}(A_p=r|A_c=m,X) \nonumber\\
     &=  \frac{1}{|\mathcal{X}_m|}\sum_{X\in\mathcal{X}_m}\hat{p}(A_p=r|X)  \label{eqn:c}
\end{align}
Let $\hat{\mathbf{C}}_{mr}$ denotes $\frac{1}{|\mathcal{X}_m|}\sum_{X\in\mathcal{X}_m}\hat{p}(A_p=r|X)$. Since the trained sensitive attribute predictor $g(\cdot)$ can be leveraged to calculate $\hat{p}(A_p|X)$, we use the trained $g(\cdot)$ to conduct inference on clean dataset $\mathcal{D}_c$ to obtain $\hat{p}(A_p=r|X)$. Then we can further calculate $\hat{\mathbf{C}}_{mr}$.
It is worth noting that the estimation accuracy of $\mathbf{C}$ depends on the following factors: (1) $\hat{p}(A_p=r|X)$ is a good estimation of $p(A_p=r|X)$; (2) The number of clean sensitive attributes, i.e., the size of $\mathcal{D}_c$; (3) The conditional Independence assumption.

\subsubsection{Learning Private Sensitive Attribute Corrector}

With the estimated sensitive attribute corruption matrix $\hat{\mathbf{C}}$, we can train a \textbf{Private Sensitive Attribute Corrector} $g'(X)$ to correct the noisy sensitive attributes. First, we multiply $\hat{\mathbf{C}}$ with the output of $g'(X)$ as the new output. Then we train $g'(X)$ on both clean data $\mathcal{D}_c$ and private data $\mathcal{D}_p$ by solving the following optimization problem:
\begin{equation}\label{eqn:glc}
   \min_{\theta_g, \theta_{g'}} \mathcal{L}_{corr}= \mathbb{E}_{(X,A_c)\in\mathcal{D}_c}\ell(A_c,g'(X))+\mathbb{E}_{(X,A_p)\in\mathcal{D}_p}\ell(A_p,\hat{\mathbf{C}}^\texttt{T}g'(X))
\end{equation}
where $\ell$ is a differentiable loss function to measure the prediction error, such as the cross-entropy loss.

\begin{algorithm}[t]%
        \caption{Training process of {\m}.}
  \label{alg:training}
  \begin{algorithmic}[1]
  
      \Require Clean data $\mathcal{D}_c=\{\mathcal{X}_c,\mathcal{A}_c, \mathcal{Y}_c\}$, private data $\mathcal{D}_p=\{\mathcal{X}_p,\mathcal{A}_p,\mathcal{Y}_p\}$
      \Ensure Learn a fair classifier $f_{\theta_Y}$
      \State // \texttt{Learning to correct private sensitive attributes}
      \State Learn a classifier $g(\cdot)$ on $\mathcal{D}_p$ \label{glc_start}
        \State Initialize corruption matrix $\hat{\mathbf{C}}\in\mathbf{R}^{l\times l}$
        \State Estimate the corruption matrix as in Eqn.\ref{eqn:c} \label{c_start} 
        \State Initialize a new classifier $g'(\cdot)$
        \State Train $g'$ with the loss function $\ell(A_c,g'(X))$ on $\mathcal{D}_c$, and the loss function $\ell(A_p,\hat{\mathbf{C}}^\texttt{T}g'(X))$ on $\mathcal{D}_p$ as in Eqn.~\ref{eqn:glc} \label{glc_end}
        \State // \texttt{Incorporate corrected sensitive attributes for semi-private  adversarial debiasing}
        \State Generate data with corrected sensitive attributes $\mathcal{D}_P^{'}=\{\mathcal{X}_p, \mathcal{A}_p^{'},\mathcal{Y}_p\}$ \label{correct}
        \State Conduct semi-private adversarial training with loss 
        function as Eqn.~\ref{eqn:ll} on $\mathcal{D}_P^{'}$ and $\mathcal{D}_c$.
        \State Return classifier $f_{\theta_Y}$ for fair prediction
  \end{algorithmic}
    
  \end{algorithm}

\subsection{{\m}: Integrating Corrected Sensitive Attributes into Semi-Private Adversarial Debiasing}
After Private Sensitive Attribute Correction, {\m} integrates the corrected sensitive attributes into semi-private adversarial debiasing for fair classification. Specifically, we leverage the Private Sensitive Attribute Corrector $g'(X)$ to construct
the set of instances with corrected sensitive attributes  $\mathcal{D}_p^{'}=\{\mathcal{X}_p, \mathcal{A}_p^{'},\mathcal{Y}_p\},\forall X\in \mathcal{X}_p$, $A_p^{'}=g'(X) \in \mathcal{A}_p^{'}$,
 which is used to optimize  $f_{\theta_p}$ as in Equation~\ref{eqn:p}.
Formally, 
the objective function of $f_{\theta_p}$ is:
  \begin{align}\label{eqn:pp} \min_{\theta_{h}}\max_{\theta_p} 
\mathcal{L}_p^{'}  &=  \mathbb{E}_{X \sim p(X \mid A_p^{'}=1)}[\log (f_{\theta_p}(h(X)))] \nonumber\\
& +\mathbb{E}_{X\sim p(X \mid A_p^{'}=0)}[\log (1-f_{\theta_p}(h(X)))] 
 \end{align}
Thus, the overall objective function of our final model {\m} is:
 \begin{align}\label{eqn:ll}
\min_{\theta_{h},\theta_Y}\max_{\theta_c,\theta_p} \mathcal{L} = \mathcal{L}_Y-\beta (\mathcal{L}_c+\alpha \mathcal{L}_p^{'})
\end{align}
We adopt the mini-batch gradient descent with Adadelta~\cite{zeiler2012adadelta} optimizer to learn the parameters.

\subsection{Privacy Discussions of {\m}}
\label{sec:Discussions_on_Privacy_Guarantee}

In this section, we illustrate the reasons \textbf{why {\m} does not impact the privacy guarantee provided by the local differential privacy mechanism in the dataset collection stage}. Based on the definition of LDP mechanism, we can infer the three key properties of LDP including \textit{Post-processing Property}, \textit{Sequential Composition Property}, and \textit{Parallel Composition Property}, which have been discussed in previous works~\cite{xiong2020comprehensive, 10.1561/0400000042,murakami2018utilityoptimized,DBLP:journals/sensors/WangZFY20}. Here we only discuss the first property of LDP as follows:

\begin{property} 
{\sc\textbf{Post-processing Property}}: If a randomized mechanism $\mathcal{M}_1$ satisfies $\epsilon$-LDP, for any randomized mechanism $\mathcal{M}_2$ (even may not satisfy LDP), then the composition of $\mathcal{M}_1$ and $\mathcal{M}_2$, namely $\mathcal{M}_2(\mathcal{M}_1(\cdot))$ also satisfies $\epsilon$-LDP.
\end{property}

The \textit{Post-processing Property} suggests that the privacy guarantee of LDP mechanism is preserved by any post-processing randomized algorithms. Machine learning models such as {\m} can be regarded as a kind of randomized algorithm. Thus, when the privacy guarantee of most of the sensitive attributes in the dataset is ensured in the data collection stage, the privacy guarantee is preserved by our proposed {\m} in the model debiasing stage.

It is worth noting that privacy has different notions including \textbf{\textit{differential privacy notion}} and 
\textbf{\textit{inference privacy notion}}
~\cite{Yu2017DynamicDL,10.1145/3436755,mireshghallah2020privacy}. The former notion only promotes privacy in the indistinguishability aspect but does not fully protect against inference attacks of adversaries using prior information, whereas the latter notion promotes privacy on the resilience against inference with prior information but does not guarantee privacy with respect to indistinguishability. Thus, these two notions are complementary~\cite{Yu2017DynamicDL,jayaraman2022attribute}. 

In this paper, we only focus on the \textit{differential privacy notion}. In our proposed semi-private setting, most of the sensitive attributes in the dataset are protected under the local differential privacy mechanism and the privacy guarantee is preserved by {\m}. Specifically, based on Lemma~\ref{lemma1}, when an attacker queries the sensitive attribute of a private instance, the probability of knowing the exact value is protected under the LDP guarantee. But the LDP mechanism does not fully protect against inference attackers with prior information, which refers to the information of non-sensitive attributes and non-private sensitive attributes in our semi-private setting.

\subsection{Fairness Discussions of {\m}}

In this section, we perform a theoretical analysis and show that under mild assumptions Equal Opportunity can be achieved when {\m} reaches the global optimum in the semi-private setting.

\begin{theorem}\label{theo2}
Let $\hat{Y}$ denote the predicted labels, $E$ denote the hidden representation of $h(X)$, $A_p^{'}$ denote the corrected sensitive attributes, $\tilde{A}_p$ denote the ground-truth sensitive attributes. If:\\
(1) The result of the private sensitive attribute correction is not totally random when the label $Y$ is positive, i.e., $p(\tilde{A}_p=1|A_p^{'}=1, Y=1)\neq p(\tilde{A}_p=1|A_p^{'}=0, Y=1)$; \\ 
(2) For all $X\in\mathcal{D}_p^{'}$,   $A_p^{'}$ and  hidden representation $E$ are conditionally independent given $\tilde{A}_p$ and the label  $Y$ is positive, i.e., $p(E,A_p^{'}|\tilde{A}_p, Y=1) = p(E|\tilde{A}_p, Y=1)p(A_p^{'}|\tilde{A}_p, Y=1)$;\\
(3) The minimax game of Equation~\ref{eqn:ll}  reaches the global optimum;\\
Then the label prediction $f_{\theta_Y}$ will achieve {\sc\textbf{Equal Opportunity}}, i.e., $p(\hat{Y}|\tilde{A}_p=0, Y=1)=p(\hat{Y}|\tilde{A}_p=1, Y=1)$ and $p(\hat{Y}|A_c=0, Y=1)=p(\hat{Y}|A_c=1, Y=1)$ {\sf (The detailed proofs can be seen in Appendix~\ref{sec:proofs})}
\end{theorem}

\section{Experiments}
\label{section_experiments}

\begin{table}
\centering 
\begin{tabular}{lccccc}
\toprule
Data & SA & \multicolumn{2}{c}{\# Train} & {\# Test}\\
& & Clean (20\%) & Private (80\%) &  & \\
\midrule
\textbf{ADULT} & Gender &  4,884 & 19,536 & 24,421 \\
\textbf{COMPAS} & Race& 611 & 2,446  & 3,058\\
\textbf{MEPS} & Race& 1,573 & 6,292 & 7,866\\
\bottomrule
\end{tabular} 
\vspace{0.1cm}
\caption{The statistics of the datasets. Clean refers to non-private sensitive attributes, whereas Private refers to the private ones with LDP. SA refers to sensitive attributes.}
\label{tab:data}
\vspace{-0.9cm}
\end{table}

\begin{table*}[!t]
    \centering
    \renewcommand\arraystretch{1.1}
    \tabcolsep=0.1cm
    \begin{tabular}{llcccccccccccc}
    \toprule
        &
        \multirow{2}{*}{\textbf{Methods}} & 
        \multicolumn{4}{c}{\textbf{ADULT}} & 
        \multicolumn{4}{c}{\textbf{COMPAS}}&
        \multicolumn{4}{c}{\textbf{MEPS}} \\
        \cmidrule(lr){3-6} \cmidrule(lr){7-10} \cmidrule(lr){11-14}
        \noalign{\vskip -0.4ex}
        && Acc. ($\uparrow$) & F1 ($\uparrow$)   & $\Delta_{DP}$ ($\downarrow$) & $\Delta_{EO}$ ($\downarrow$)  
        & Acc. ($\uparrow$) & F1 ($\uparrow$)   & $\Delta_{DP}$ ($\downarrow$) & $\Delta_{EO}$ ($\downarrow$)  
        & Acc. ($\uparrow$) & F1 ($\uparrow$)   & $\Delta_{DP}$ ($\downarrow$) & $\Delta_{EO}$ ($\downarrow$) 
        \\ 
        \cline{1-14}
        \multirow{10}{*}{\rotatebox[origin=r]{90}{\textbf{$\epsilon$ = 0.5}}}
        &\textbf{Vanilla} 
        &84.8$\pm$0.2 &65.4$\pm$0.7 &9.1$\pm$0.4 &5.3$\pm$1.0 
        &67.0$\pm$0.6 &64.3$\pm$0.9 &13.8$\pm$1.1 &12.8$\pm$1.4
        &86.1$\pm$0.8 &48.5$\pm$1.5 &4.5$\pm$0.5 &4.5$\pm$1.0 \\
        &\textbf{RemoveS}$^{\dagger}$ 
        &84.9$\pm$0.3 &64.8$\pm$0.8 &8.4$\pm$0.2 &4.1$\pm$1.1  
        &67.3$\pm$0.8 &64.2$\pm$1.2 &13.0$\pm$0.4 &12.2$\pm$0.6 
        &86.1$\pm$0.2 &49.9$\pm$1.6 &4.7$\pm$0.5 &4.6$\pm$1.1 \\
        \cdashline{2-14}
        
        &\textbf{RNF-GT}~\cite{du2021fairness}
        &83.5$\pm$1.2 &63.3$\pm$0.8 &9.0$\pm$1.1 &5.1$\pm$0.5 
        &66.9$\pm$0.8 &63.5$\pm$0.9 &13.9$\pm$0.6 &13.1$\pm$1.3 
        &85.8$\pm$0.1 &49.5$\pm$1.5 &5.0$\pm$0.3 &4.9$\pm$0.9 \\
        &\textbf{FairRF}~\cite{zhao2021fair}$^{\dagger}$
        &84.0$\pm$0.5 &63.5$\pm$0.7 &8.2$\pm$0.3 &3.6$\pm$0.8 
        &66.3$\pm$0.7 &63.2$\pm$0.5 &13.8$\pm$2.4 &13.5$\pm$1.2 
        &85.9$\pm$0.2 &47.0$\pm$1.9 &4.9$\pm$1.0 &4.7$\pm$1.3 \\
        &\textbf{CorScale}~\cite{lamy2019noise}
        &84.1$\pm$0.3 &63.4$\pm$0.3 &8.1$\pm$0.9 &3.5$\pm$0.5 
        &66.8$\pm$0.5 &63.9$\pm$0.2 &13.5$\pm$1.3 &12.6$\pm$0.8 
        &85.6$\pm$0.4 &47.3$\pm$1.4 &4.6$\pm$1.2 &4.5$\pm$0.9 \\
        \cdashline{2-14}
        &\textbf{Clean}$^{\dagger}$
        &84.9$\pm$0.4 &64.6$\pm$0.7 &8.4$\pm$0.4 &4.1$\pm$1.0 
        &67.2$\pm$0.6 &64.8$\pm$1.0 &13.1$\pm$0.5 &12.3$\pm$0.8 
        &86.1$\pm$0.1 &50.6$\pm$1.6 &4.8$\pm$0.6 &4.4$\pm$1.2 \\
        &\textbf{Private}
        &84.7$\pm$0.3 &64.6$\pm$0.3 &8.4$\pm$0.3 &4.1$\pm$1.2 
        &67.1$\pm$0.7 &64.6$\pm$1.1 &13.0$\pm$0.4 &12.1$\pm$0.7
        &86.0$\pm$0.1 &50.8$\pm$2.3 &4.8$\pm$0.7 &4.5$\pm$1.1 \\
        &\textbf{C+P}
        &84.8$\pm$0.5 &64.8$\pm$0.6 &8.1$\pm$0.2 &3.4$\pm$1.4
        &67.2$\pm$0.6 &63.9$\pm$1.1 &12.9$\pm$0.2 &12.2$\pm$0.5 
        &86.1$\pm$0.1 &48.8$\pm$1.8 &4.4$\pm$0.4 &4.3$\pm$0.7 \\
        \cline{2-14}
        \noalign{\vskip 0.25ex}
        &\textbf{{\m}}
        &84.7$\pm$0.4 &64.5$\pm$0.7 &\cellcolor{gray!20}\textbf{7.8}$\pm$0.3 &\cellcolor{gray!20}\textbf{2.3}$\pm$1.2 
        &67.0$\pm$1.6 &63.8$\pm$1.4 &\cellcolor{gray!20}\textbf{12.7}$\pm$0.5 &\cellcolor{gray!20}\textbf{12.1}$\pm$0.6 
        &86.0$\pm$0.1 &47.3$\pm$1.7 &\cellcolor{gray!20}\textbf{4.1}$\pm$0.8 &\cellcolor{gray!20}\textbf{4.0}$\pm$1.2 \\

        \noalign{\vskip 0.2ex}
        \cline{1-14}
        \multirow{10}{*}{\rotatebox[origin=r]{90}{\textbf{$\epsilon$ = 1}}}
        &\textbf{Vanilla} 
        &84.9$\pm$0.3 &65.6$\pm$0.4 &9.3$\pm$0.2 &5.3$\pm$0.5 
        &67.2$\pm$0.5 &64.4$\pm$0.5 &13.9$\pm$0.7 &12.9$\pm$1.0
        &86.5$\pm$0.2 &48.6$\pm$0.6 &4.6$\pm$0.4 &4.7$\pm$0.8 \\
        &\textbf{RemoveS}$^{\dagger}$ 
        &84.9$\pm$0.3 &64.8$\pm$0.8 &8.4$\pm$0.2 &4.1$\pm$1.1  
        &67.3$\pm$0.8 &64.2$\pm$1.2 &13.0$\pm$0.4 &12.2$\pm$0.6 
        &86.1$\pm$0.2 &49.9$\pm$1.6 &4.7$\pm$0.5 &4.6$\pm$1.1 \\
        \cdashline{2-14}
        
        &\textbf{RNF-GT}~\cite{du2021fairness}
        &83.4$\pm$0.8 &63.1$\pm$0.5 &8.8$\pm$1.2 &4.9$\pm$0.4 
        &66.8$\pm$0.5 &63.4$\pm$0.4 &13.7$\pm$0.5 &12.8$\pm$1.1 
        &85.6$\pm$0.2 &49.4$\pm$1.1 &4.8$\pm$0.2 &4.8$\pm$0.7 \\
        &\textbf{FairRF}~\cite{zhao2021fair}$^{\dagger}$
        &84.0$\pm$0.5 &63.5$\pm$0.7 &8.2$\pm$0.3 &3.6$\pm$0.8 
        &66.3$\pm$0.7 &63.2$\pm$0.5 &13.8$\pm$2.4 &13.5$\pm$1.2 
        &85.9$\pm$0.2 &47.0$\pm$1.9 &4.9$\pm$1.0 &4.7$\pm$1.3 \\
        &\textbf{CorScale}~\cite{lamy2019noise}
        &84.1$\pm$0.4 &63.8$\pm$0.4 &8.1$\pm$0.1 &3.3$\pm$0.2 
        &66.5$\pm$0.5 &63.6$\pm$0.3 &13.3$\pm$1.2 &12.2$\pm$0.7 
        &85.5$\pm$0.3 &47.2$\pm$1.3 &4.5$\pm$0.8 &4.0$\pm$0.9 \\
        \cdashline{2-14}
        &\textbf{Clean}$^{\dagger}$
        &84.9$\pm$0.4 &64.6$\pm$0.7 &8.4$\pm$0.4 &4.1$\pm$1.0 
        &67.2$\pm$0.6 &64.8$\pm$1.0 &13.1$\pm$0.5 &12.3$\pm$0.8 
        &86.1$\pm$0.1 &50.6$\pm$1.6 &4.8$\pm$0.6 &4.4$\pm$1.2 \\
        &\textbf{Private}
        &84.4$\pm$0.6 &63.9$\pm$0.5 &8.2$\pm$0.2 &3.6$\pm$0.9 
        &67.4$\pm$0.1 &64.5$\pm$0.9 &12.9$\pm$0.8 &12.1$\pm$0.4
        &86.3$\pm$0.3 &49.9$\pm$1.7 &4.3$\pm$0.5 &4.3$\pm$0.7 \\
        &\textbf{C+P}
        &84.5$\pm$0.3 &64.4$\pm$0.3 &8.0$\pm$0.4 &3.2$\pm$0.4
        &67.5$\pm$0.6 &63.3$\pm$1.4 &12.8$\pm$0.4 &12.2$\pm$0.3 
        &86.0$\pm$0.2 &47.5$\pm$1.0 &4.0$\pm$0.3 &4.1$\pm$0.2 \\
        
        \cline{2-14}
        \noalign{\vskip 0.25ex}
        
        &\textbf{{\m}}
        &84.6$\pm$0.4 & 64.1$\pm$0.7 &\cellcolor{gray!20}\textbf{7.5}$\pm$0.1 &\cellcolor{gray!20}\textbf{1.3}$\pm$1.4 
        &66.9$\pm$1.0 &64.0$\pm$0.9
        &\cellcolor{gray!20}\textbf{12.5}$\pm$0.7 &\cellcolor{gray!20}\textbf{11.7}$\pm$0.9 
        &85.8$\pm$0.2 &44.4$\pm$1.9
        &\cellcolor{gray!20}\textbf{3.2}$\pm$0.5 &\cellcolor{gray!20}\textbf{2.3}$\pm$0.7 \\
        \noalign
        {\vskip -0.2ex}
    \bottomrule
    \end{tabular}
    \vspace{0.2cm}
    \caption{The performance comparison for fair classification under the semi-private setting. $^{\dagger}$: The performance does not change when the privacy budget \textbf{$\epsilon$} varies because local differential privacy (LDP) is only enforced on sensitive attributes following previous works~\cite{lamy2019noise} and \textit{FairRF}, \textit{RemoveS} do not need sensitive attributes, \textit{Clean} only utilizes the very limited clean ones.}
    \vspace{-0.5cm}
    \label{tab:performance}
\end{table*}

In this section, we conduct experiments to evaluate the performance of our method and try to answer the following research questions: 
\textbf{RQ1}: Can {\m} obtain fair predictions with mostly  private sensitive attributes? 
\textbf{RQ2}: What is the impact of the amount of data with clean sensitive attributes?
\textbf{RQ3}: How do different privacy budgets impact fair classification performance?
\textbf{RQ4}: How does the private sensitive attribute correction affect the performance of prediction and fairness?

\subsection{Experimental Settings}\label{sec:setting}

\subsubsection{Datasets}\label{sec:data}
We conduct experiments on three typical datasets for fair classification and statistics are shown in Table~\ref{tab:data}: 
\begin{itemize}
    \item \textbf{COMPAS}\cite{Julia2016machine}: This dataset describes the task of predicting the recidivism of individuals with \textbf{race} as the sensitive attribute. 
    \item \textbf{ADULT}\cite{asuncion2007uci}: This dataset is utilized to predict whether an individual's income exceeds 50k with \textbf{gender} as the sensitive attribute. 
    \item \textbf{MEPS}\cite{cohen2003design}: 
    The dataset is to predict one's utilization in a set of large-scale medical surveys with \textbf{race} as the sensitive attribute.
\end{itemize}

\subsubsection{Evaluation Setting}\label{sec:data}
We report results on the test set 
and all experiments are repeated 5 times. The average result and standard deviation are reported in Table~\ref{tab:performance}. For each run, the  dataset is randomly divided into train set and test set. The set of random  seeds for five runs is \{5, 7, 11, 19, 29\}. We compare {\m} and the baseline at different values of the {\em clean ratio}, which is defined as:
\begin{equation}\label{eqn:cd}
    \text{clean ratio}=\frac{\text{\#samples w/ clean SA}}{\text{ \#samples w/ clean SA + \#samples w/ private SA}}
\end{equation}
\subsubsection{Baselines}
    (1) \textbf{Vanilla}: A vanilla MLP classifier without any debiasing method.;
    (2) \textbf{RemoveS}: Directly removing the sensitive attributes in the input;
    (3) \textbf{RNF-GT}~\cite{du2021fairness}: A recently proposed state-of-the-art debiasing method utilizing mixup for debiasing that requires the sensitive attribute annotations;
    (4) \textbf{FairRF}~\cite{zhao2021fair}: A debiasing method that does not need any sensitive attributes;
    (5) \textbf{CorScale}~\cite{lamy2019noise}: A debiasing method that regards all sensitive attributes as noisy ones;
    (6) \textbf{Clean}: An adversarial debiasing model on only the instances with clean sensitive attributes;
    (7) \textbf{Private}: An adversarial debiasing model on only the instances with private sensitive attributes;
    (8) \textbf{C+P} (Clean+Private): We simply merge both sets (treating the private sensitive attributes as the clean ones) and use them together for training an adversarial debiasing model.

\subsubsection{Evaluation Metrics}
\label{sec:Evaluation_Metrics}
Following existing work on fair classification, we measure the classification performance with Accuracy (Acc.) and F1, and the fairness performance based on \textit{Demographic Parity} and \textit{Equal Opportunity}~\cite{mehrabi2021survey}.

\begin{itemize}
    \item Demographic Parity: 
    it requires each demographic group has the same chance for a positive outcome: 
$\mathbb{E}(\hat{Y}|A=1)=\mathbb{E}(\hat{Y}|A=0)$. 
We report the difference of each  group's demographic parity: $\Delta_{DP}=|\mathbb{E}(\hat{Y}|A=1)-\mathbb{E}(\hat{Y}|A=0)|$

\item Equal Opportunity: 
it requires the true positive rate of different groups is equal: $\mathbb{E}(\hat{Y}|A=1, Y=1)=\mathbb{E}(\hat{Y}|A=0, Y=1)$. 
We report the difference of each sensitive group's equal opportunity: $\Delta_{EO}=|\mathbb{E}(\hat{Y}|A=1, Y=1)-\mathbb{E}(\hat{Y}|A=0, Y=1)|$
\end{itemize}
Note that demographic parity and equal opportunity measure fairness performance in different ways. The fairness performance is better with smaller  values of $\Delta_{EO}$ and $\Delta_{DP}$.

\subsection{Main Result Analysis}
\label{sec:Performance_of_Fair_Classification}

To answer \textbf{RQ1}, we compare {\m} with baselines on three benchmark datasets for fair classification. For each experiment, we select five random seeds to partition the original dataset into a training set and a test set. We randomly select 80\% of the training set as the private part and the others as the non-private part (i.e., clean ratio =20\%). For the private samples, we set the privacy budget $\epsilon$ as 0.5 and 1, which means the flipping probability on sensitive attributes is around 37.75\% and 26.89\% respectively according to the Lemma~\ref{lemma1}. The average performance and standard deviation over five times are reported in Table ~\ref{tab:performance}. We have the following observations:

\begin{figure*}[htbp!]
\vspace{-0.15cm}
	\centering
 	\subfigure[Accuracy]{
 	\includegraphics[width=0.24\textwidth]{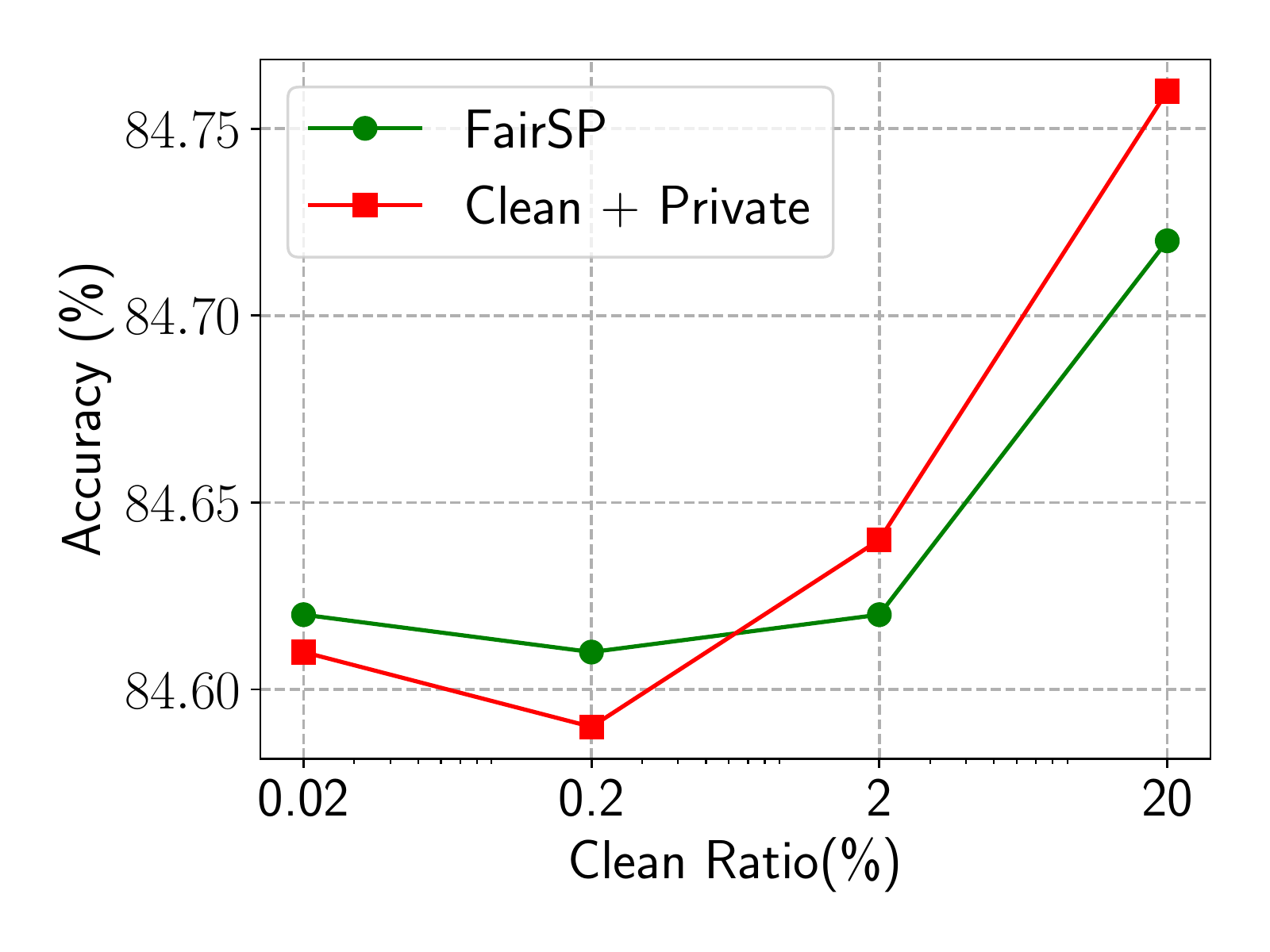}}
		\subfigure[F1]{
	\includegraphics[width=0.24\textwidth]{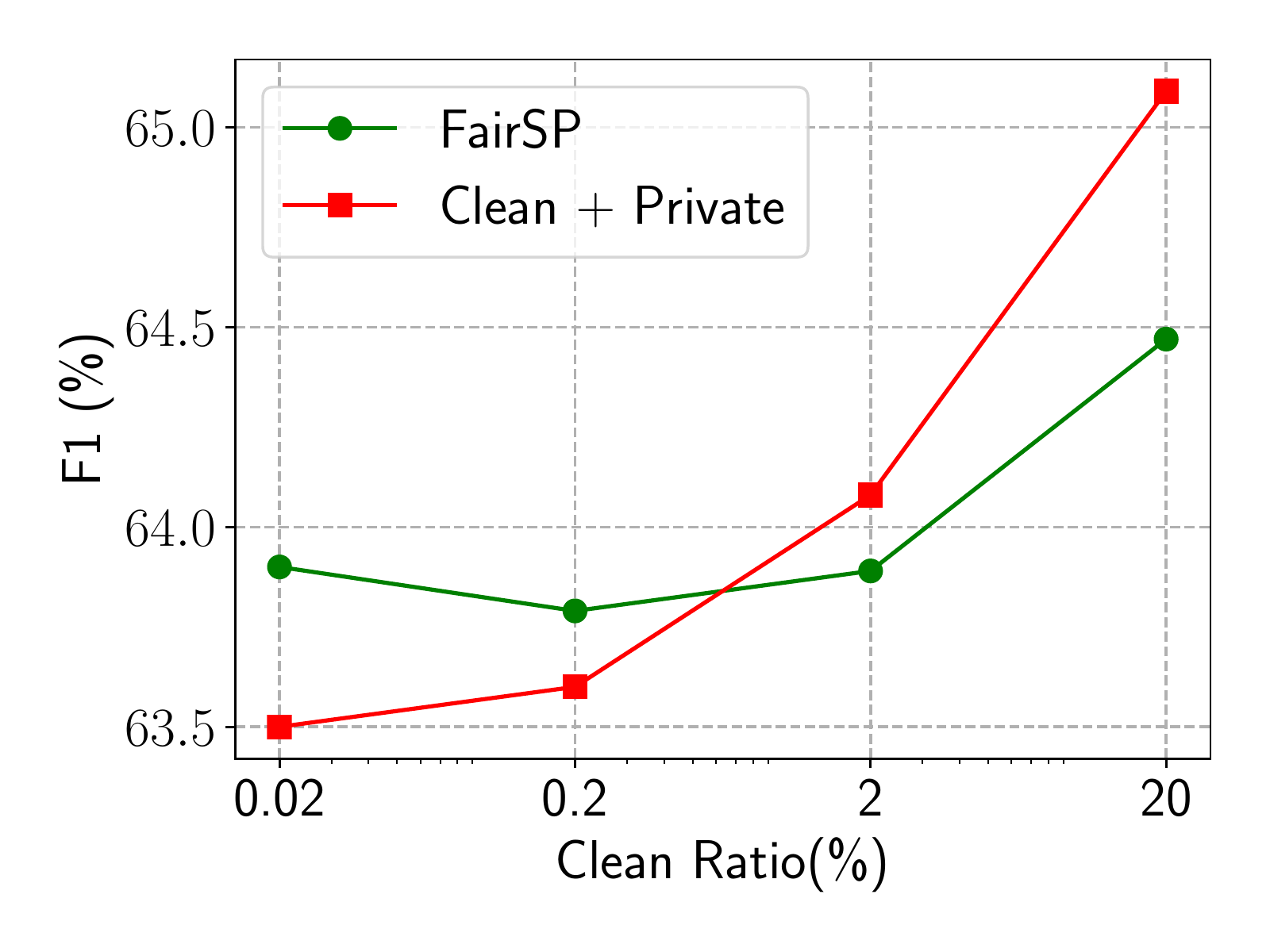}}
		\subfigure[$\Delta_{DP}$]{
	\includegraphics[width=0.24\textwidth]{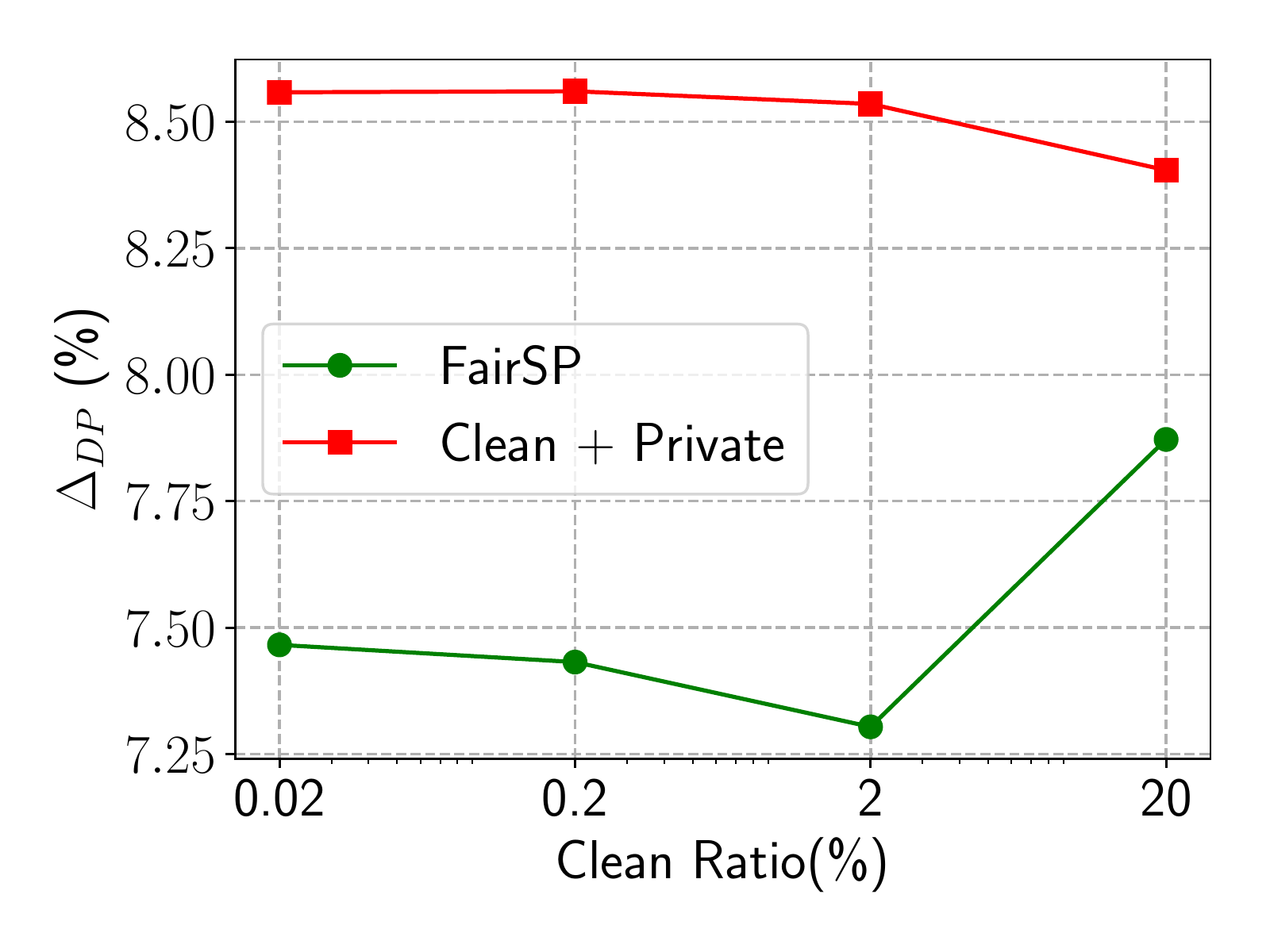}}
			\subfigure[$\Delta_{EO}$]{
	\includegraphics[width=0.24\textwidth]{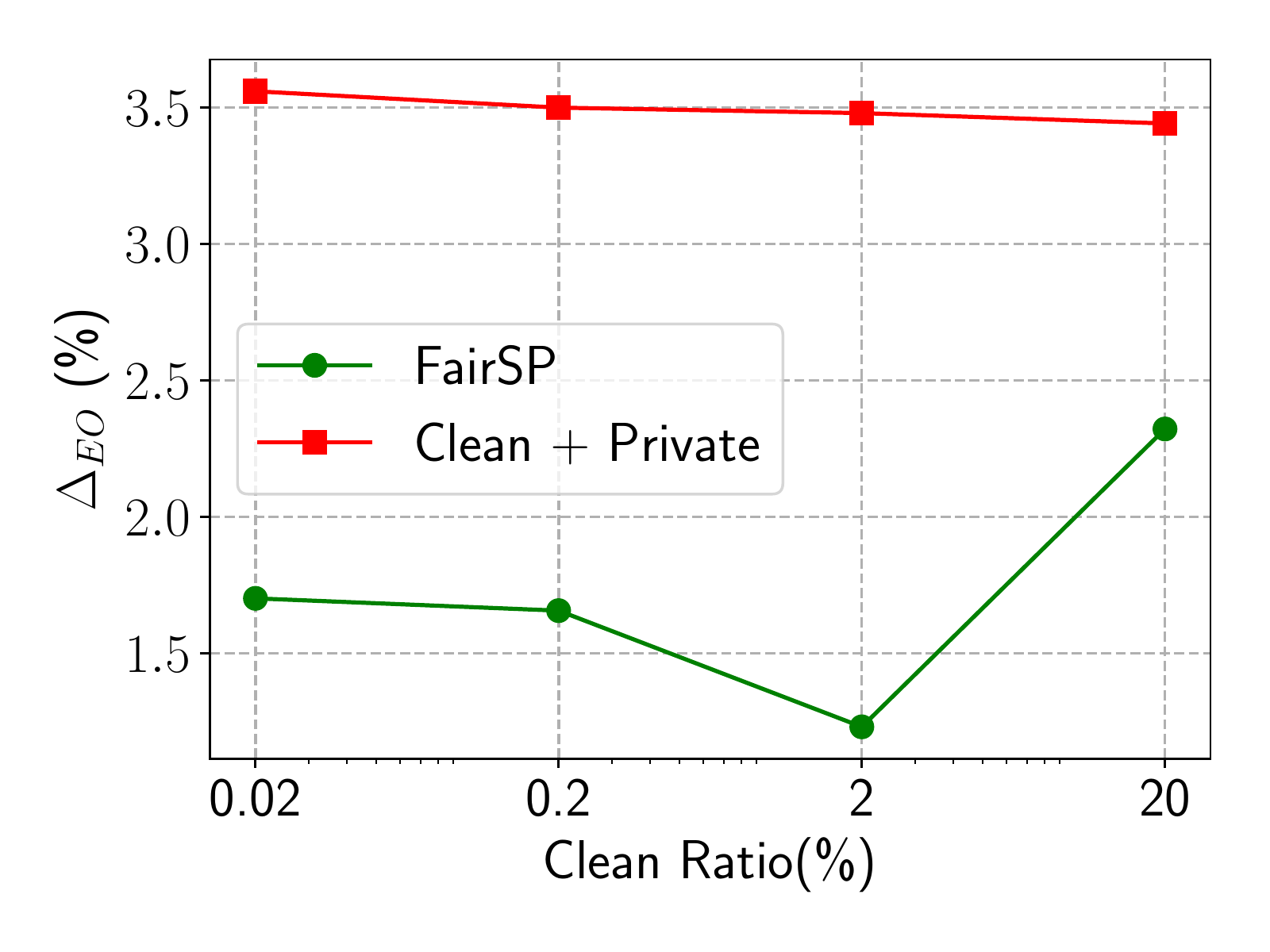}}
 	\vspace{-0.25cm}
	 \caption{The impact of clean data ratio on prediction and debiasing performances  on ADULT.
	 }
	\label{fig:private_ratio}
	\vspace{-0.2cm}
\end{figure*}

\begin{figure*}[htbp!]
\vspace{-0.15cm}
	\centering
	\subfigure[Accuracy]{
	\includegraphics[width=0.24\textwidth]{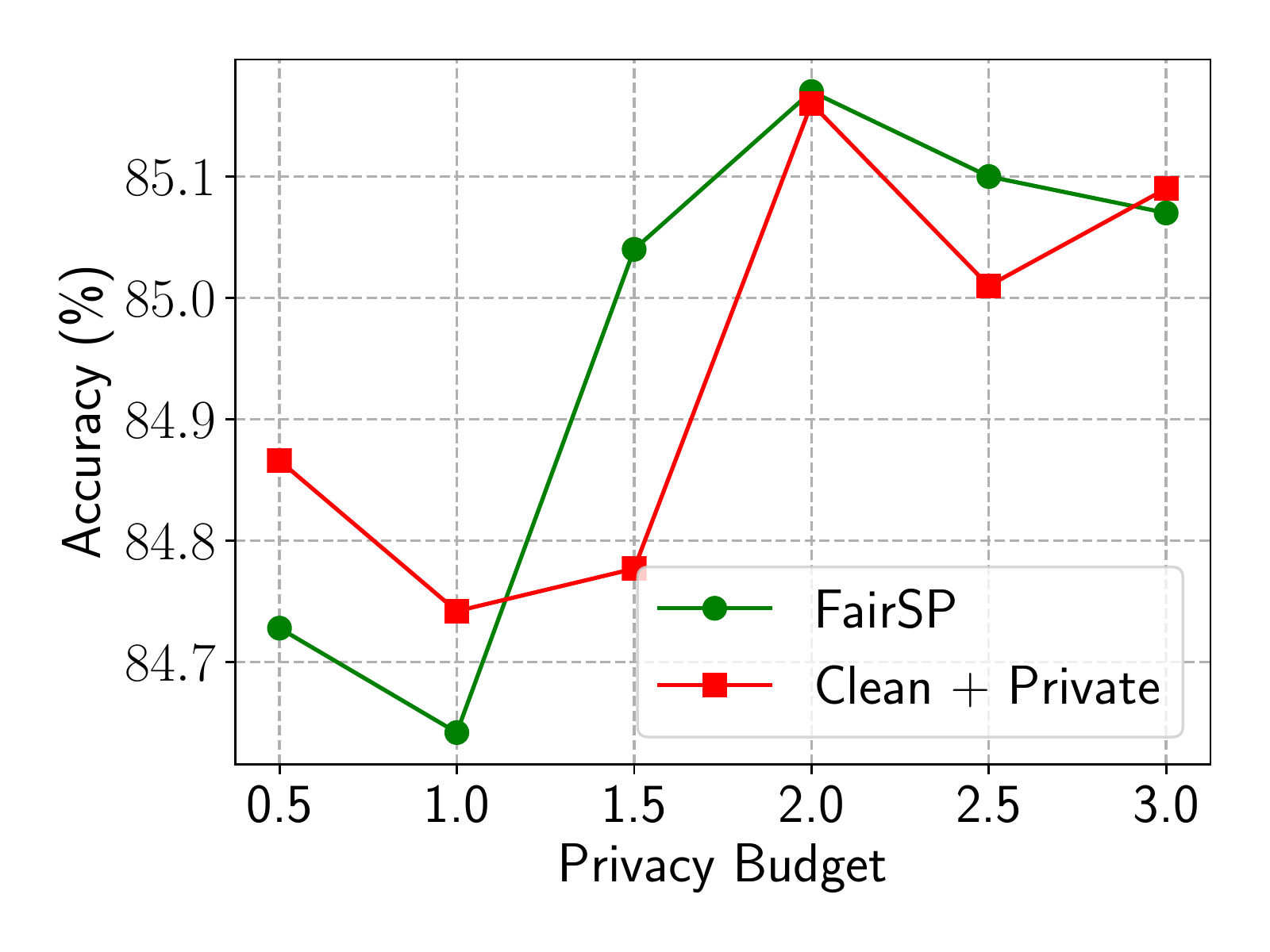}}
		\subfigure[F1]{
	\includegraphics[width=0.24\textwidth]{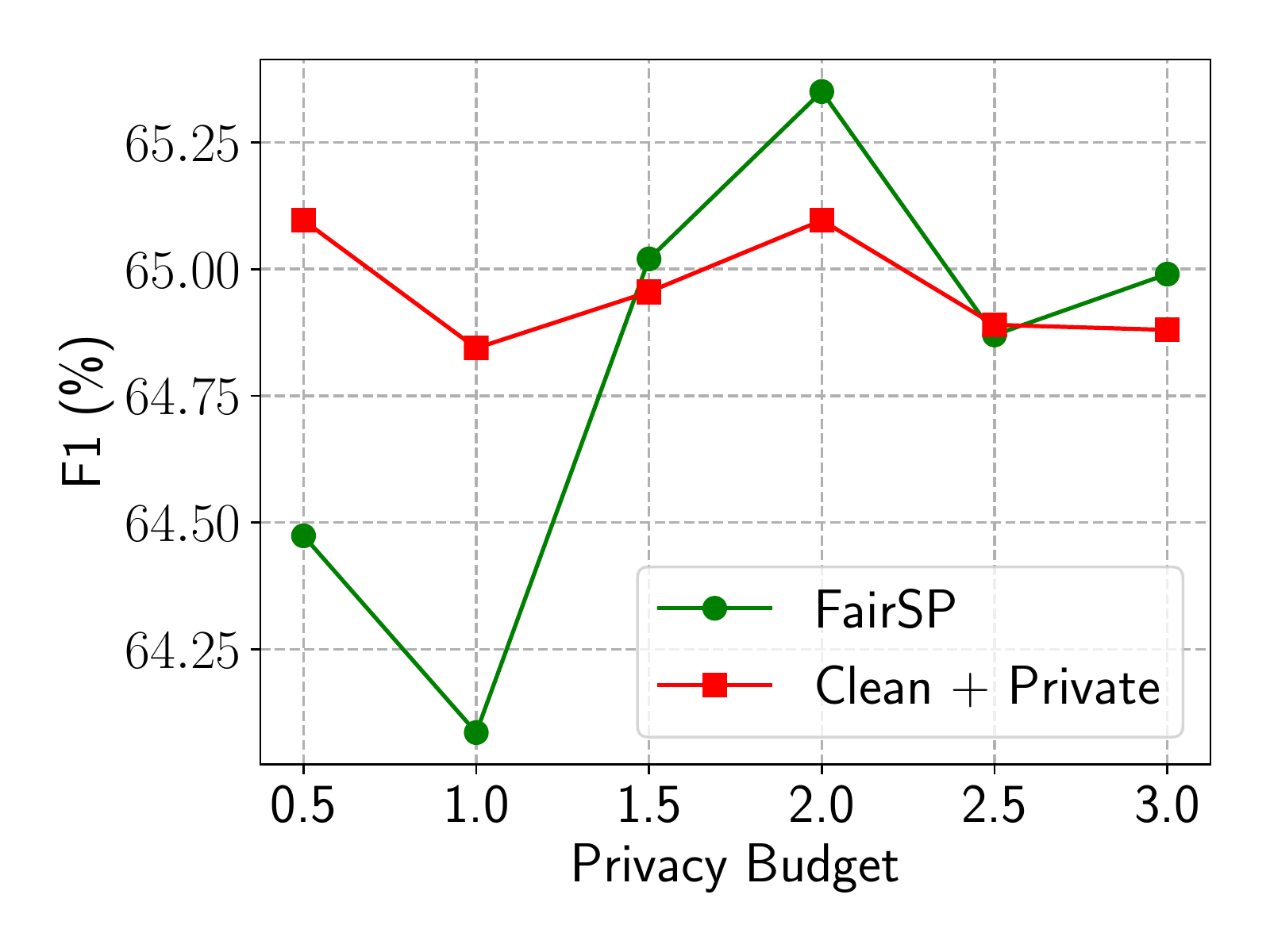}}
		\subfigure[$\Delta_{DP}$]{
	\includegraphics[width=0.24\textwidth]{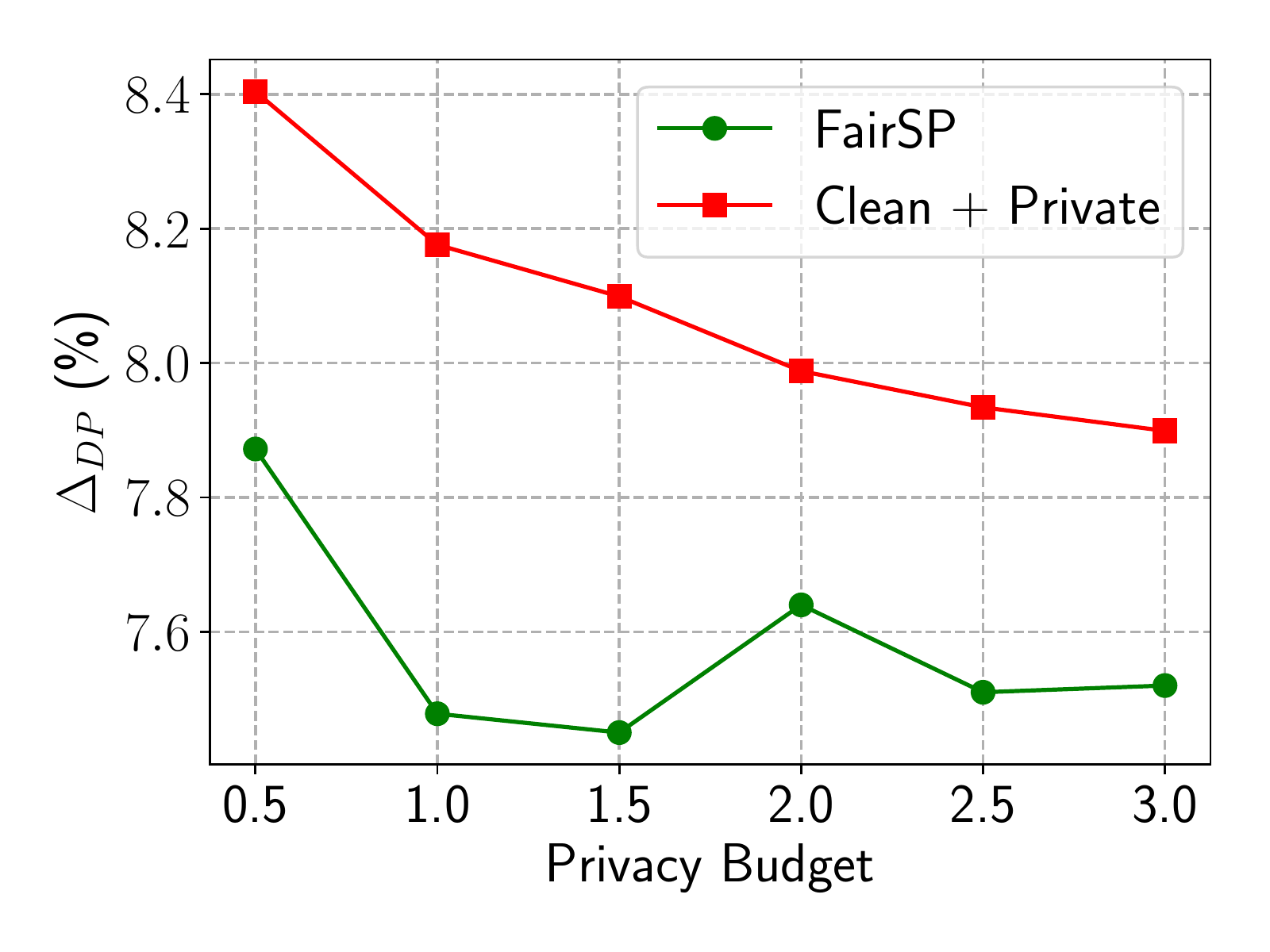}}
			\subfigure[$\Delta_{EO}$]{
	\includegraphics[width=0.24\textwidth]{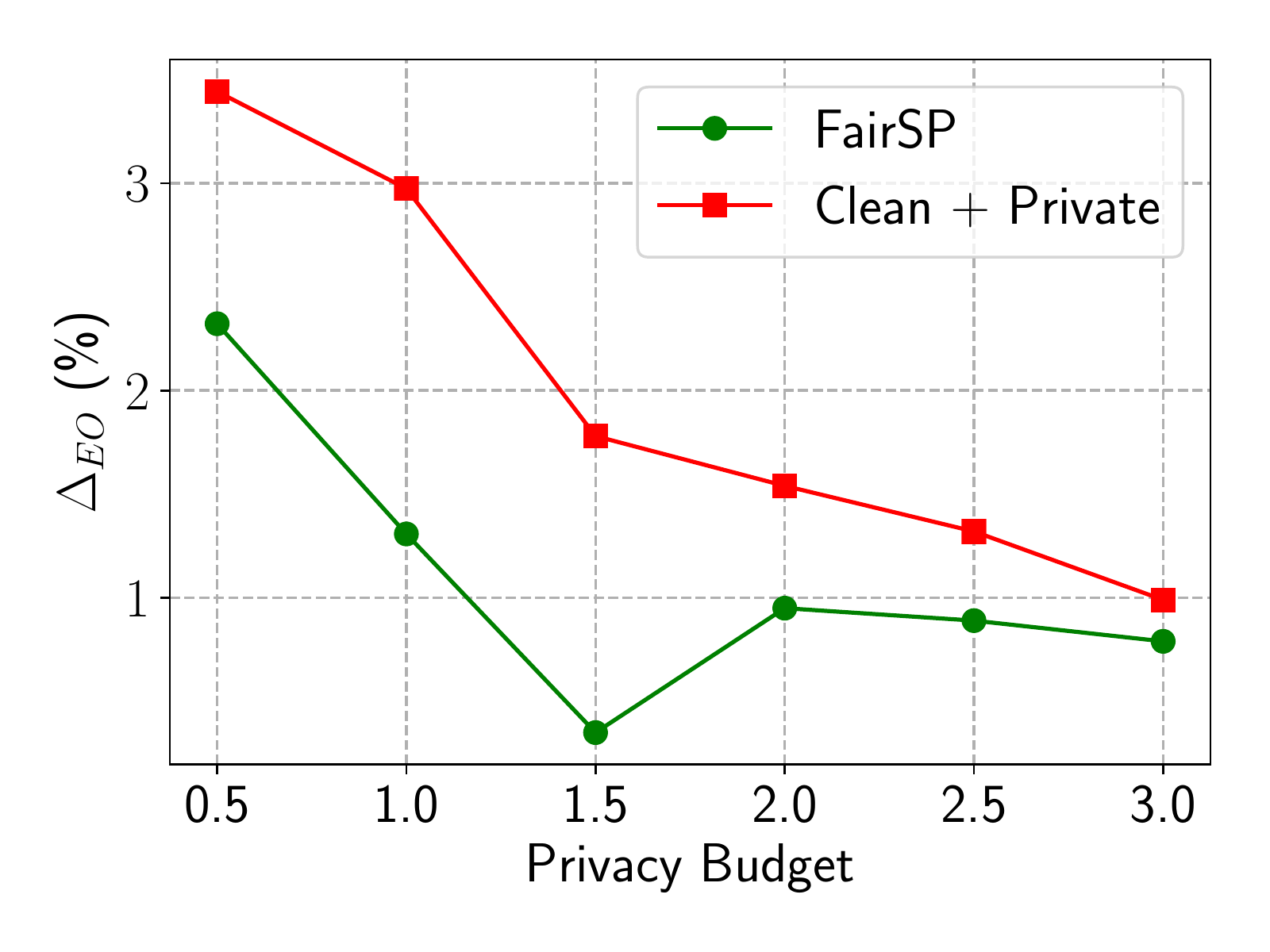}}
 	\vspace{-0.25cm}
	\caption{The impact of privacy budget $\epsilon$ on prediction and debiasing performances on ADULT.
	}\label{fig:private_budget}
	\vspace{-0.4cm}
\end{figure*}

\begin{itemize}
    \item In general, we observe that \textbf{{\m} can achieve the best fairness performance without causing a significant drop in prediction performance} under the semi-private setting on three datasets, compared to other baselines. 
    For example, compared with the recently proposed state-of-the-art debiasing model RNF-GT, {\m} has achieved  54.90\% improvement in terms of $\Delta_{EO}$ and 13.33\% over $\Delta_{DP}$ on ADULT when privacy budget $\epsilon = 0.5$. 
    
    \item We observe that \textbf{conventional debiasing methods that directly leverage sensitive attributes are generally ineffective} under the semi-private scenario. For example, comparing RNF-GT with the Vanilla model, we can observe that their fairness performances are similar w.r.t. $\Delta_{DP}$ and $\Delta_{EO}$ on three datasets.
    
    \item We can see that \textbf{it is important to leverage the very limited clean sensitive attributes} to ensure fairness in the semi-private scenario. For example, {\m} has consistent improvement on $\Delta_{DP}$ and $\Delta_{EO}$ compared with FairRF and CorScale, which do not leverage the clean sensitive attributes.
    \item \textbf{Exploiting  the limited clean sensitive attributes and private ones jointly is important} for debiasing in the semi-private setting. 
    We can generally observe that {\m} $>$ Clean+Private $>$ Clean $\approx$ Private  $\approx$ RemoveS $>$  Vanilla for debiasing performances. 
    First, {\m} and  ``Clean+Private'' perform better than the other three baselines, which shows that leveraging both clean and private sensitive attributes is necessary. 
    Second, the observation that ``Clean'', ``Private'', RemoveS perform similarly indicates that only relying on clean or noisy data is less effective for debiasing.

\end{itemize}

\subsection{Additional Analysis}
\label{sec:Additional_Analysis}
\subsubsection{Impact of Private Data Ratio}
\label{sec:Private_Data_Ratio}

In this subsection, we investigate the impact of different private data ratios, to answer  \textbf{RQ2}. 
We conduct four groups of experiments with different clean data ratios (defined as in Eqn.~\ref{eqn:cd}) in the range of [0.02\%, 0.2\%, 2\%, 20\%] on the ADULT dataset. Each group of experiments has the same privacy budget $\epsilon$ as  0.5. For each clean data ratio, we compare our proposed model {\m} with ``Clean+Private'' and demonstrate the prediction metric Accuracy, F1 as well as fairness metric $\Delta_{DP}$ and $\Delta_{EO}$. For each experiment, we run five times and report the average result in Figure~\ref{fig:private_ratio}. From the figure, we can make the following observations:

\begin{itemize}
    \item In general, we can observe from Figure~\ref{fig:private_ratio} (c) (d) that our proposed \textbf{{\m} has a consistent improvement in fairness performance by a large margin} compared with the baseline ``Clean+Private'' consistently at different clean data ratios. 
     \item \textbf{Even with extreme small clean data ratio such as 0.02\%, {\m} maintains relatively stable fairness performances} on ADULT dataset. This indicates Private Sensitive Attribute Correction is still effective with extremely limited clean data.
    \item From Figure~\ref{fig:private_ratio} (a) (b), we observe that the proposed \textbf{{\m} demonstrates comparable prediction performance with the baseline}  regardless of different clean data ratios. 
\end{itemize}

\subsubsection{Impact of Privacy Budget}
To answer \textbf{RQ3}, in this subsection, we investigate the impact of different privacy budgets on fair classification performances.  We conduct six groups of experiments with different privacy budget  $\epsilon$ as \{0.5, 1.0, 1.5, 2.0, 2.5, 3.0\}  on the ADULT dataset, which means the flipping probability on sensitive attributes is around \{38\%, 27\%, 18\%, 12\%, 7\%, 4\%\} respectively. The clean data ratio for each group of experiments is 20\%. For each Privacy Budget  $\epsilon$, we compare our proposed model {\m} and baseline ``Clean+Private'' with performance metrics Accuracy, F1, and fairness metrics $\Delta_{DP}$ and $\Delta_{EO}$. 
From Figure~\ref{fig:private_budget}, we have the following observations:

\begin{itemize}
    \item In general, we can observe from  Figure~\ref{fig:private_budget} (c) (d) that our proposed model \textbf{{\m} performs consistently better compared to  ``Clean+Private''}. With a larger privacy budget  $\epsilon$, the fairness performances of both models are improved. This is because the dataset has fewer noisy sensitive attributes under larger  $\epsilon$.
    \item \textbf{{\m} is more effective when the privacy guarantee is strong}. With a larger privacy budget  $\epsilon$,  the gap between {\m} and ``Clean+Private'' is  narrower. This may be because the estimation of the private sensitive attributes in {\m} becomes less effective when the flipping probability is smaller.
    \item From  Figure~\ref{fig:private_budget} (a) and (b), we see that   \textbf{{\m} has comparable classification performances} on Accuracy and F1 with the baseline ``Clean+Private'' regardless of privacy budget $\epsilon$. 
    
\end{itemize}

\subsubsection{Ablation Study of Private Sensitive Attribute Correction}

\begin{figure}[tbp!]
    \centering

      	\subfigure[Comparison on F1]{
	\includegraphics[width=0.23\textwidth]{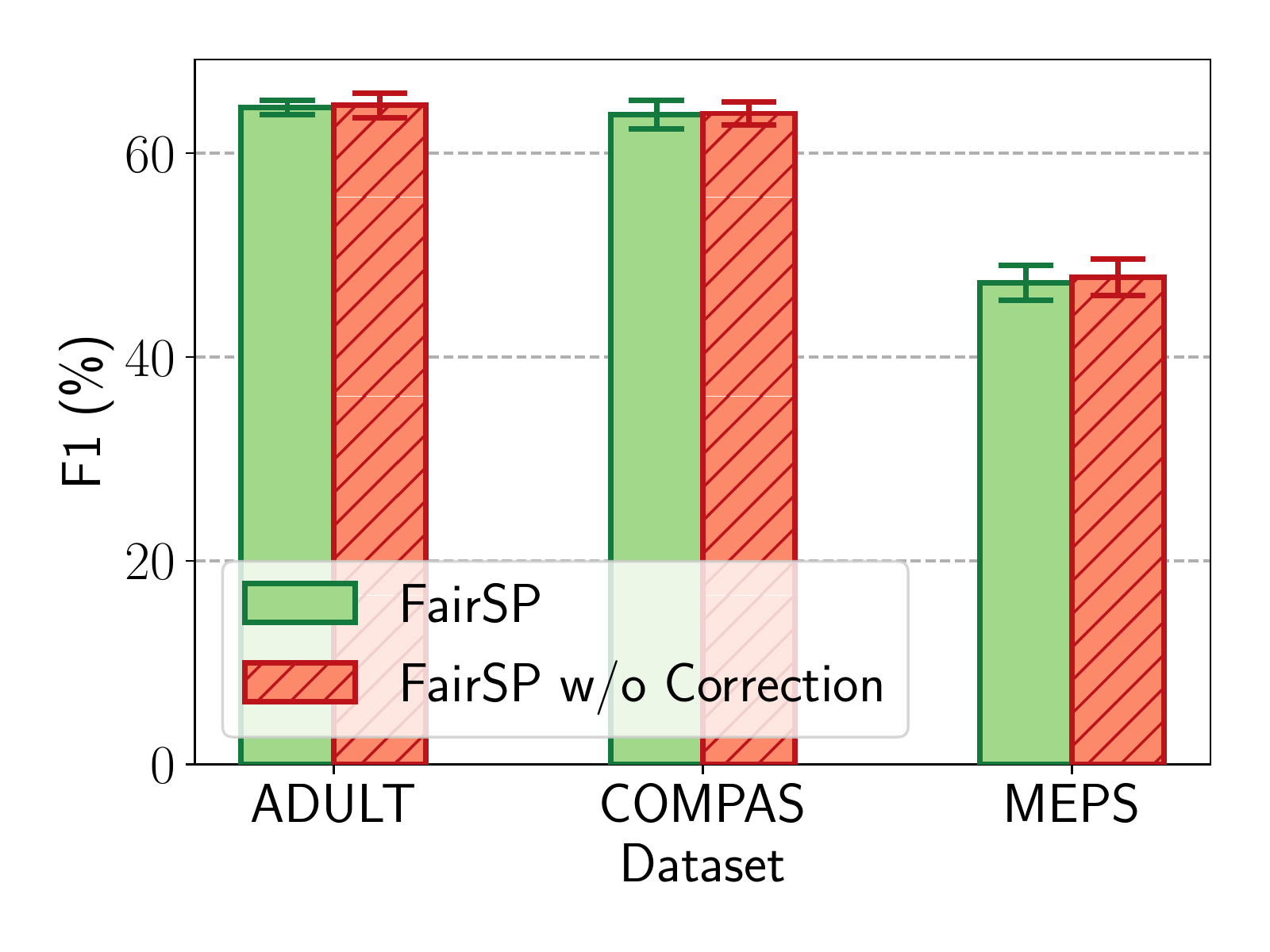}}
		\subfigure[Comparison on $\Delta_{EO}$]{
	\includegraphics[width=0.23\textwidth]{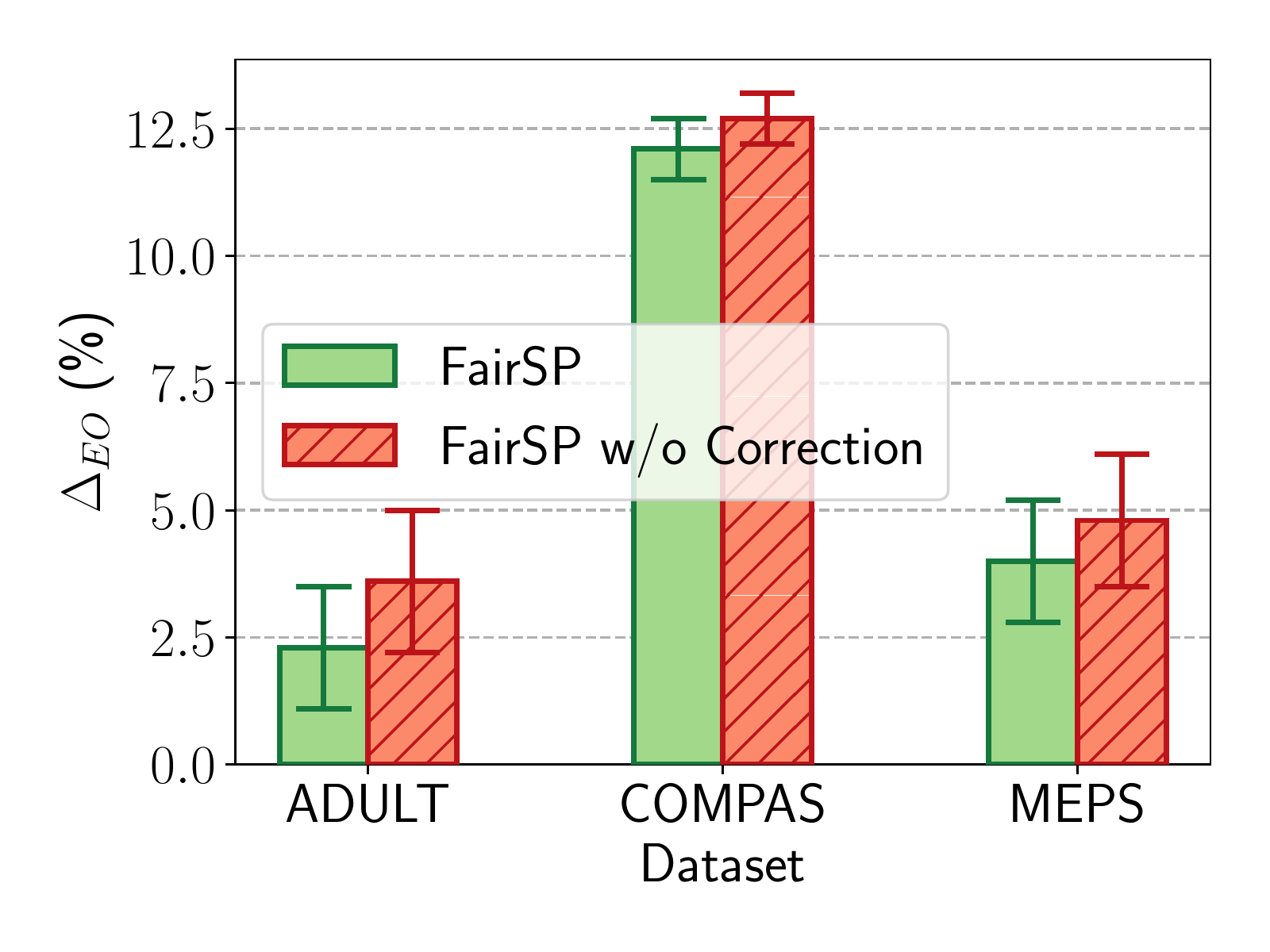}}
    
	\vspace{-0.3cm}
    
    \caption{Imapct of private sensitive attribute correction.  }
    \label{fig:ablation}
	\vspace{-0.5cm}
\end{figure}

Now we investigate the impact of private sensitive attribute correction, to answer \textbf{RQ4}. 
We keep the setting of the training set and test set division the same as the main results in Table~\ref{tab:performance}, and show the results on ADULT, COMPAS, and MEPS datasets when privacy budget $\epsilon = 0.5$.
The average performance and standard deviation for five rounds are reported in Figure~\ref{fig:ablation}. We can make the following observations: (1) The proposed  correction strategy on private sensitive attributes is consistently effective in improving debiasing performance  on three datasets and has the largest improvement over ADULT. (2) The proposed correction method does not cause a significant drop in prediction performance over three datasets.

\subsubsection{Parameter Sensitivity Analysis}
\begin{figure}[tbp!]
\hspace{-0.15cm}
	\centering
	\subfigure[Accuracy]{
	\includegraphics[width=0.23\textwidth]{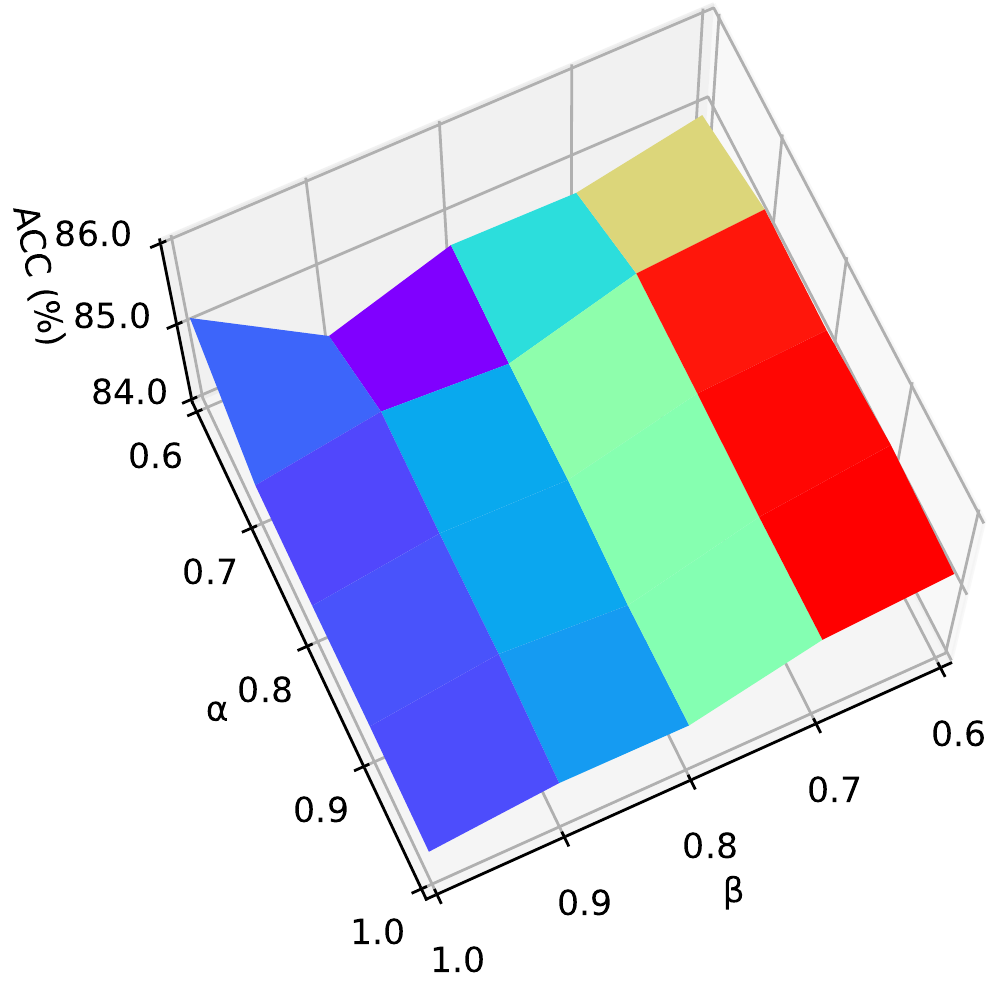}}
	\hspace{-0.3cm}
		\subfigure[F1]{
	\includegraphics[width=0.23\textwidth]{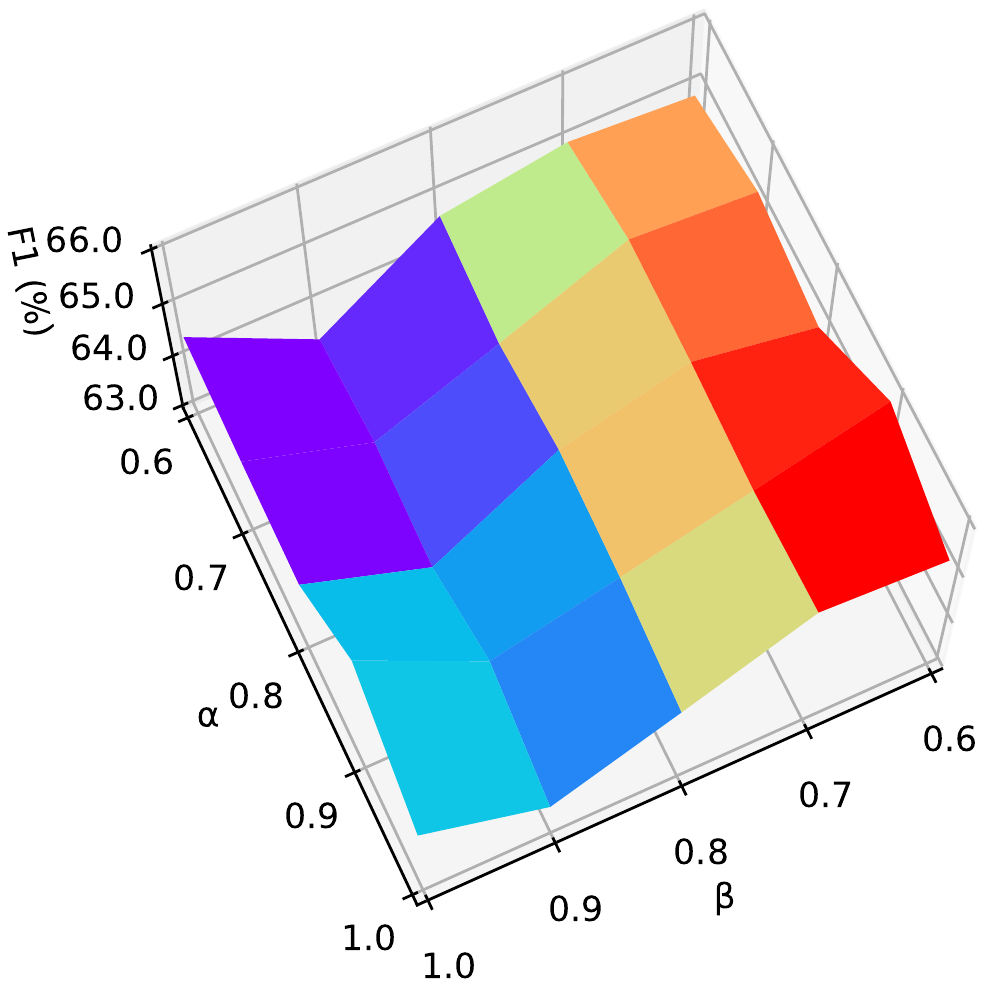}}
		\subfigure[$\Delta_{DP}$]{
	\includegraphics[width=0.23\textwidth]{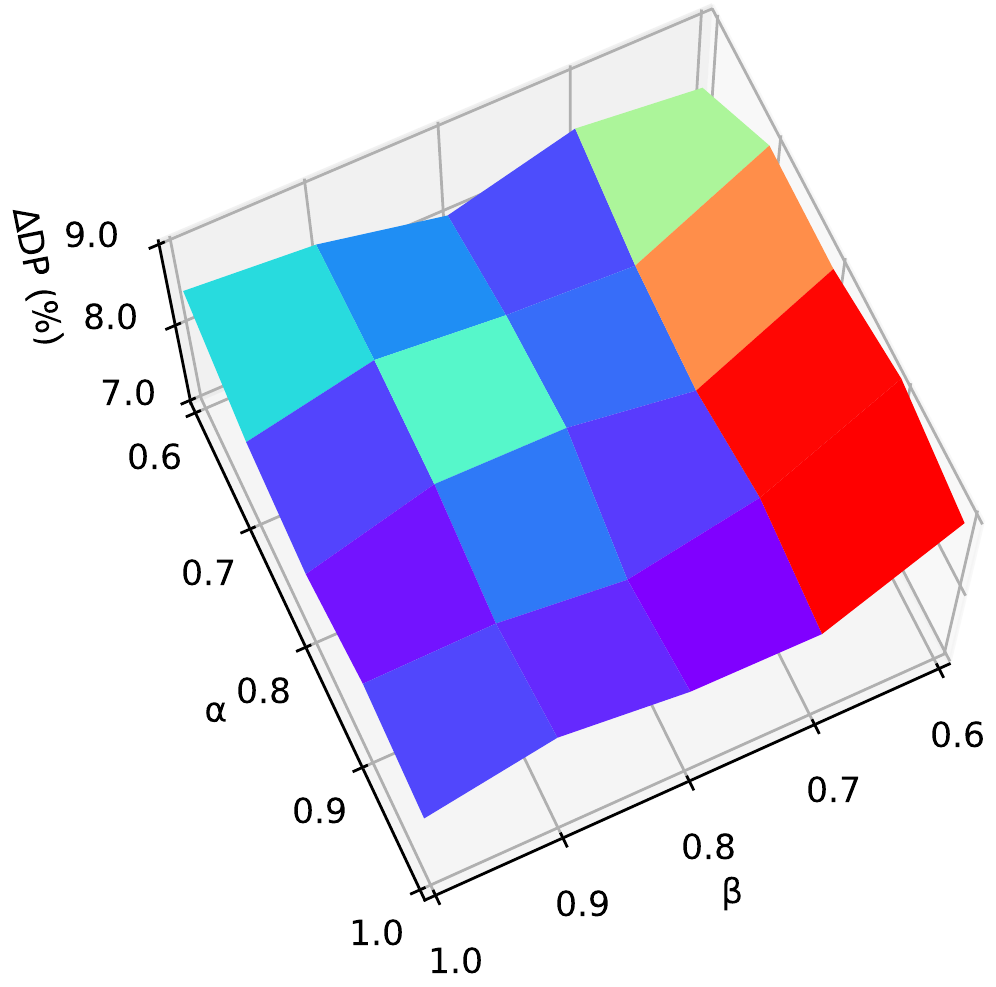}}
	\hspace{-0.28cm}
			\subfigure[$\Delta_{EO}$]{
	\includegraphics[width=0.23\textwidth]{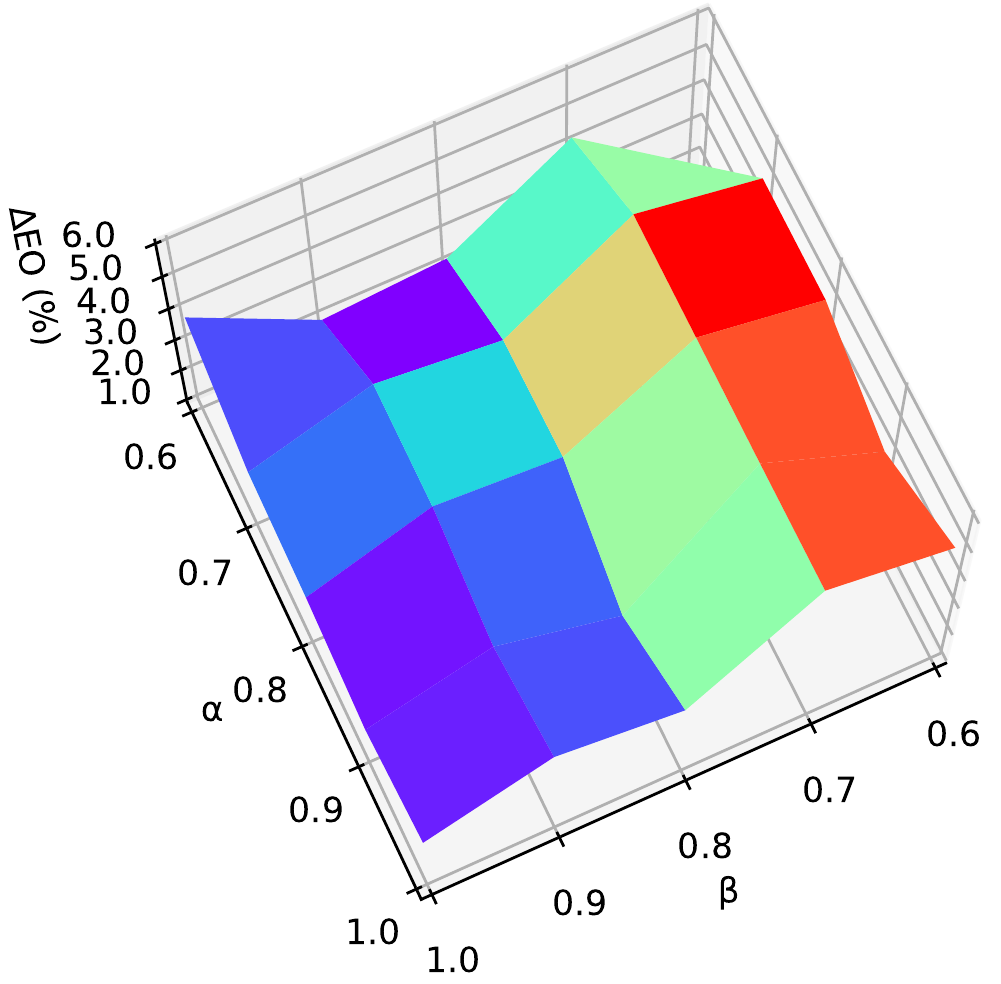}}
	\caption{The parameter sensitive analysis of {\m}.}\label{fig:parameter}
	\vspace{-0.35cm}
\end{figure}

We now explore the parameter sensitivity of the two important hyperparameters of our model: $\alpha$ controls the impact of the adversarial private sensitive attribute predictor, while $\beta$ controls the influence of the adversary on debiasing. We vary $\alpha$ in [0.6, 0.7, 0.8, 0.9, 1.0] and $\beta$ from [0.6, 0.7, 0.8, 0.9, 1.0]. As shown in Fig.~\ref{fig:parameter} (the value increases from purple to red), we can observe that: (1) The performances of Accuracy and F1 are relatively consistent in the range, and the trends of $\Delta_{DP}$ and $\Delta_{EO}$ are also similar. (2) When  $\beta$ is larger, the prediction performance drops  and fairness performance is  improved, which demonstrates the trade-off between fairness and prediction performances. (3) Based on the experiments, we can achieve optimal  fairness and comparable accuracy when selecting both $\alpha$ and $\beta$ as 1.0.

\section{Related Work}
\label{section_relatedwork}
In this section, we briefly describe the related work on (1) Fairness in machine learning; and (2) Differential privacy in machine learning.

\textbf{Fairness in Machine Learning}
Recent research on fairness in machine learning has drawn significant attention to develop effective algorithms to achieve fairness and maintain good prediction performance. Existing methods generally focus on individual fairness~\cite{kang2020inform,cheng2021socially,DBLP:journals/corr/abs-2010-04053} or group fairness~\cite{hardt2016equality,zhang2017achieving}.
Other niche notions of fairness include subgroup fairness~\cite{kearns2018preventing} and Max-Min fairness\cite{lahoti2020fairness}.
The majority of existing debiasing techniques have been applied at different stages of a machine learning model~\cite{mehrabi2021survey} including  \textit{pre-processing}~\cite{kamiran2012data},
\textit{in-processing}~\cite{agarwal2018reductions,bechavod2017learning} 
and \textit{post-processing} approaches~\cite{dwork2018decoupled}.
Such machine learning methods generally require the access to sensitive attributes, which is often infeasible in practice. 
Very few recent works study fairness with limited sensitive attributes available or without sensitive attributes~\cite{dai2021say,lahoti2020fairness,zhao2021fair}. For example, Dai \textit{et al.} propose to achieve fairness on graph neural networks when the sensitive attributes are limited~\cite{dai2021say}.

\textbf{Differential Privacy in Machine Learning.}
Differential privacy (DP)~\cite{10.1561/0400000042} is a widely adopted approach to
provide strong privacy guarantee regardless of the adversaries' prior knowledge~\cite{mcsherry2007mechanism}, which can protect user privacy in various machine learning tasks including supervised learning and unsupervised learning~\cite{ji2014differential,gong2020survey,abadi2016deep,vaidya2013differentially}. 
Recently, local differential privacy (LDP) has been extensively studied in the distributed setting such that private data can be locally perturbed without a trusted aggregator~\cite{cormode2018privacy,wang2019answering}. 
Due to the inherent connection of privacy and fairness (e.g., protecting or debiasing on user sensitive attributes), several recent works look into the trade-offs and mutual risks between privacy and fairness~\cite{agarwal2021trade,chang2021privacy}. 
Other approaches aim to ensure both DP and fairness while preserving good utility~\cite{xu2019achieving}, or learn fair models with only private data~\cite{mozannar2020fair,lamy2019noise,wang2020robust}, which are different from our semi-private setting.

\section{Conclusion and Future Work}
\label{section_conclution}
In this paper, we study a novel problem of fair classification with semi-private sensitive attributes. We develop an end-to-end adversarial debiasing model {\m} to jointly learn from  very limited clean sensitive attributes and mostly private ones under local differential privacy (LDP) definition. We provide a theoretical analysis of the conditions that fairness can be achieved under the semi-private setting.
Extensive experimental results on real-world datasets 
demonstrate the effectiveness of the proposed framework.
For future work, first, we will study fairness in a more general semi-private setting on a variety of data types such as text and graphs and analyze the fairness and privacy guarantee on these data. 
Second, we will investigate fairness under more privacy mechanisms such as federated learning, anonymization, and encryption.

\newpage
\clearpage
\bibliographystyle{ACM-Reference-Format}
\bibliography{main_2}

\newpage 
\clearpage
\appendix

\setcounter{theorem}{0}
\setcounter{lemma}{0}
\setcounter{definition}{0}
\setcounter{proposition}{0}
\setcounter{property}{0}

\begin{table}[t]
\centering \caption{Notation Table.}
\begin{tabular}{ll}
\toprule
Notation & Meaning \\
\midrule
$\mathcal{D}$ & the whole set of instances\\
$\mathcal{D}_c$ & set of instances with clean sensitive attributes \\
$\mathcal{D}_p$& set of instances with private sensitive attributes \\
$\mathcal{D}_p^{'}$ & set of instances with corrected sensitive attributes\\
$\mathcal{A}$ & the whole set of sensitive attributes \\
$\mathcal{A}_c$ & set of clean sensitive attributes \\
$\mathcal{A}_p$& set of  private sensitive attributes \\
$\mathcal{A}_p^{'}$ & set of corrected sensitive attributes\\
$\mathcal{X}$ & the whole set of samples excluding \\&sensitive attributes \\
$\mathcal{X}_c$ & set of samples in $\mathcal{D}_c$  excluding \\&sensitive attributes \\
$\mathcal{X}_p$& set of samples in $\mathcal{D}_p$  excluding \\&sensitive attributes \\
$\mathcal{X}_m$ & subset of $\mathcal{X}_c$ with the corresponding \\&sensitive attribute as $m$ \\
$\mathcal{Y}$ & the whole set of label \\
$\mathcal{Y}_c$ & set of label in $\mathcal{D}_c$ \\
$\mathcal{Y}_p$& set of label in $\mathcal{D}_p$ \\
\midrule
$A$ & sensitive attribute \\
$A_c$ & clean sensitive attribute \\
$A_p$& private sensitive attribute \\
$A_p^{'}$ & corrected sensitive attribute\\
$\tilde{A}_p$ & ground-truth sensitive attribute of $A_p$\\
$X$ & sample excluding sensitive attributes\\
$Y$ & label\\
$\hat{Y}$ & predicted result of $f_{\theta_Y}$\\
\midrule
 $\mathbf{C}$& sensitive attribute corruption matrix \\
  $\mathbf{C}_{mr}$& the corruption possibility from $A_c=m$ to $A_p=r$ \\
  $\hat{\mathbf{C}}$& the estimation of sensitive attribute corruption \\ &matrix \\
  $\hat{\mathbf{C}}_{mr}$& the estimation of the corruption possibility \\ &from $A_c=m$ to $A_p=r$ \\
   $g$ & function for estimating sensitive attribute  \\&corruption matrix\\
  $g'$ & function for correcting private sensitive attributes\\
 \midrule
  $h$& shared encoder for learning latent representations \\
  $f_{\theta_Y}$ & function for predicting labels\\
 $f_{\theta_c}$ & function for predicting clean sensitive attributes\\
 $f_{\theta_p}$ & function for predicting corrected sensitive \\&attributes\\
  \midrule
  $\theta_h$& set of the parameters of $h$ \\
  $\theta_Y$ & set of the parameters of $f_{\theta_Y}$ \\
 $\theta_c$ & set of the parameters of $f_{\theta_c}$ \\
 $\theta_p$ & set of the parameters of  $f_{\theta_p}$  \\
\bottomrule
\end{tabular} \label{tab:symbol}
\end{table}

\section{Summary of Notations}
A summary of the notations in this paper is in Table~\ref{tab:symbol}.

\section{Proofs}
\label{sec:proofs}

\begin{theorem}\label{theo0}
Let $E$ denote the hidden representation of $h(X)$. The global optimum of the minimax game of Equation~\ref{eqn:ll} can be achieved if and only if $p(E|A_c=1)=p(E|A_c=0)$ and $p(E|A_p^{'}=1)=p(E|A_p^{'}=0)$. 
\end{theorem}

\begin{proof}According to the Proposition 1 in~\cite{goodfellow2014generative}, the optimal adversarial sensitive attribute predictor $f_{\theta_c}$ and $f_{\theta_p}$ are as follows: 
$$f^*_{\theta_c}=\frac{p(E|A_c=1)}{p(E|A_c=1)+p(E|A_c=0)}$$ $$f^*_{\theta_p}=\frac{p(E|A_p^{'}=1)}{p(E|A_p^{'}=1)+p(E|A_p^{'}=0)}$$ 
Then the minimax game of Equation~\ref{eqn:ll}  can be written as follows (R is the objective function):
\begin{align}
R &= \mathbb{E}_{E \sim p(E \mid A_c=1)}[\log \frac{p(E|A_c=1)}{p(E|A_c=1)+p(E|A_c=0)}] \nonumber\\\nonumber
&+\mathbb{E}_{E\sim p(E \mid A_c=0)}[\log \frac{p(E|A_c=0)}{p(E|A_c=1)+p(E|A_c=0)}] \\\nonumber
&+ \alpha \mathbb{E}_{E \sim p(E \mid A_p^{'}=1)}[\log \frac{p(E|A_p^{'}=1)}{p(E|A_p^{'}=1)+p(E|A_p^{'}=0)}] \\
&+\alpha \mathbb{E}_{E\sim p(E \mid A_p^{'}=0)}[\log \frac{p(E|A_p^{'}=0)}{p(E|A_p^{'}=1)+p(E|A_p^{'}=0)}]
\end{align}

On the one hand, Since  $\alpha$ is a non-negative hyperparameter, if $p(E|A_c=1)=p(E|A_c=0)$ and $p(E|A_p^{'}=1)=p(E|A_p^{'}=0)$ exist,  the objective function $R$ can reach the minimum value $-(1+\alpha)\log(4)$. 

On the other hand, the objective function $R$ can also be formulated as:
\begin{align}
R &= -(1+\alpha)\log(4) \nonumber\\\nonumber
&+ KL\Big(p(E|A_c=1)||\frac{p(E|A_c=1)+p(E|A_c=0)}{2}\Big) \\\nonumber
&+ KL\Big(p(E|A_c=0)||\frac{p(E|A_c=1)+p(E|A_c=0)}{2}\Big)\\\nonumber
&+ \alpha\cdot KL\Big(p(E|A_p^{'}=1)||\frac{p(E|A_p^{'}=1)+p(E|A_p^{'}=0)}{2}\Big)\\
&+ \alpha\cdot KL\Big(p(E|A_p^{'}=0)||\frac{p(E|A_p^{'}=1)+p(E|A_p^{'}=0)}{2}\Big)\\
&=-(1+\alpha)\log(4)+2\cdot JSD(p(E|A_c=1)||p(E|A_c=0))\nonumber\\
&+2\alpha\cdot JSD(p(E|A_p^{'}=1)||p(E|A_p^{'}=0))
\end{align}
where $KL$ denotes Kullback–Leibler divergence, and $JSD$ denotes Jensen–Shannon divergence.
Since  $\alpha$ is a non-negative hyperparameter and the Jensen–Shannon divergence between two distributions is always non-negative. And $JSD$ is zero only when they are equal. Thus, $R$ can reach the minimum value $-(1+\alpha)\log(4)$ only if $p(E|A_p^{'}=1)=p(E|A_p^{'}=0)$ and $p(E|A_c=1)=p(E|A_c=0)$. Concluding the proof.
\end{proof}%

\begin{theorem}\label{theo1}
Let $\hat{Y}$ denote the predicted labels, $E$ denote the hidden representation of $h(X)$, $A_p^{'}$ denote the corrected sensitive attributes, $\tilde{A}_p$ denote the ground-truth sensitive attributes.  If:\\
(1) The result of the private sensitive attribute correction is not totally random, i.e., $p(\tilde{A}_p=1|A_p^{'}=1)\neq p(\tilde{A}_p=1|A_p^{'}=0)$;  \\ 
(2) For all $X\in\mathcal{D}_p^{'}$,  $A_p^{'}$ and  hidden representation $E$ are conditionally independent given $\tilde{A}_p$, i.e., $p(E,A_p^{'}|\tilde{A}_p) = p(E|\tilde{A}_p)p(A_p^{'}|\tilde{A}_p)$;\\
(3) The minimax game of Equation~\ref{eqn:ll}  reaches the global optimum;\\
Then the label prediction $f_{\theta_Y}$ will achieve {\sc \textbf{Demographic Parity}}, i.e., $p(\hat{Y}|\tilde{A}_p=0)=p(\hat{Y}|\tilde{A}_p=1)$ and $p(\hat{Y}|A_c=0)=p(\hat{Y}|A_c=1)$
\end{theorem}

\begin{proof}
We first illustrate that the first two assumptions generally hold: 

(1) 
Considering there exists a correlation between non-sensitive attributes and sensitive attributes, it is reasonable to assume that our \textit{learning to correct} method does not produce totally random estimation results when it converges, i.e.,  $p(\tilde{A}_p=1|A_p^{'}=1)\neq p(\tilde{A}_p=1|A_p^{'}=0)$.

(2) Since we use the objective function in $\mathcal{L}_{corr}$ to derive the corrected sensitive attributes $A_p^{'}$, and learn the latent presentation $E$ with the objective function $\mathcal{L}$, between which they do not share any parameters, it generally holds that $A_p^{'}$ is independent with the representation of $E$, i.e., $p(E,A_p^{'}|\tilde{A}_p) = p(E|\tilde{A}_p)p(A_p^{'}|\tilde{A}_p)$; 

 We then prove Theorem~\ref{theo1} as follows: since  the assumption (2) $p(E,A_p^{'}|\tilde{A}_p) = p(E|\tilde{A}_p)p(A_p^{'}|\tilde{A}_p)$  holds, we have $(E\bot A_p^{'}) | \tilde{A}_p$, then we can have $p(E|A_p^{'},\tilde{A}_p)=p(E|\tilde{A}_p)$. 
 
 From Theorem~\ref{theo0}, we know when the our  $\mathcal{L}$ of the minimax game of Equation~\ref{eqn:ll} reaches the global optimum, we have $p(E|A_p^{'}=1)=p(E|A_p^{'}=0)$, which is equivalent with  $\sum_{\tilde{A}_p}p(E, \tilde{A}_p|A_p^{'}=1)=\sum_{\tilde{A}_p}p(E, \tilde{A}_p|A_p^{'}=0)$.
 Therefore,
$
\sum_{\tilde{A}_p}p(E|\tilde{A}_p,A_p^{'}=1)p(\tilde{A}_p|A_p^{'}=1)=\sum_{\tilde{A}_p}p(E|\tilde{A}_p,A_p^{'}=0)p(\tilde{A}_p|A_p^{'}=0)\nonumber
$
Also, $p(E|A_p^{'},\tilde{A}_p)=p(E|\tilde{A}_p)$, thus we can obtain, 
\begin{align}
\sum_{\tilde{A}_p}p(E|\tilde{A}_p)p(\tilde{A}_p|A_p^{'}=1)=\sum_{\tilde{A}_p}p(E|\tilde{A}_p)p(\tilde{A}_p|A_p^{'}=0)
\end{align}
Based on the above equation and assumption (1) $p(A_p^{'}=1|\tilde{A}_p=1)\neq p(A_p^{'}=1|\tilde{A}_p=0)$, we can get,
\begin{align}
  \frac{p(E|\tilde{A}_p=1)}{p(E|\tilde{A}_p=0)}  &=\frac{p(\tilde{A}_p=0|A_p^{'}=1)-p(\tilde{A}_p=0|A_p^{'}=0)}{p(\tilde{A}_p=1|A_p^{'}=0)-p(\tilde{A}_p=1|A_p^{'}=1)} \nonumber\\ 
  &=\frac{(1 - p(\tilde{A}_p=1|A_p^{'}=1)) - (1 - p(\tilde{A}_p=1|A_p^{'}=0))}{p(\tilde{A}_p=1|A_p^{'}=0)-p(\tilde{A}_p=1|A_p^{'}=1)}
  \nonumber\\
  &=1
\end{align}
Since we already proof that $p(E|A_c=1)=p(E|A_c=0)$ from Theorem~\ref{theo0}, and $\hat{Y}=f_{\theta_Y}(E)$, we can get $p(\hat{Y}|\tilde{A}_p=1)=p(\hat{Y}|\tilde{A}_p=0)$ and $p(\hat{Y}|A_c=1)=p(\hat{Y}|A_c=0)$, which is the {\sc \textbf{Demographic Parity}}. Thus, {\sc Demographic Parity} is achieved when the minimax game of Equation~\ref{eqn:ll} converges to the global optimum. Concluding the proof.
\end{proof}

\begin{theorem}\label{theo2}
Let $\hat{Y}$ denote the predicted labels, $E$ denote the hidden representation of $h(X)$, $A_p^{'}$ denote the corrected sensitive attributes, $\tilde{A}_p$ denote the ground-truth sensitive attributes. If:\\
(1) The result of the private sensitive attribute correction is not totally random when the label  $Y$ is positive, i.e., $p(\tilde{A}_p=1|A_p^{'}=1, Y=1)\neq p(\tilde{A}_p=1|A_p^{'}=0, Y=1)$; \\ 
(2) For all $X\in\mathcal{D}_p^{'}$,   $A_p^{'}$ and  hidden representation $E$ are conditionally independent given $\tilde{A}_p$ and the label $Y$ is positive, i.e., $p(E,A_p^{'}|\tilde{A}_p, Y=1) = p(E|\tilde{A}_p, Y=1)p(A_p^{'}|\tilde{A}_p, Y=1)$;\\
(3) The minimax game of Equation~\ref{eqn:ll}  reaches the global optimum;\\
Then the label prediction $f_{\theta_Y}$ will achieve {\sc\textbf{Equal Opportunity}}, i.e., $p(\hat{Y}|\tilde{A}_p=0, Y=1)=p(\hat{Y}|\tilde{A}_p=1, Y=1)$ and $p(\hat{Y}|A_c=0, Y=1)=p(\hat{Y}|A_c=1, Y=1)$
\end{theorem}

\begin{proof}
We first illustrate that the first two assumptions generally hold: 

(1) This assumption is the relaxation of the  assumption(1) of Theorem~\ref{theo1}, i.e., $p(\tilde{A}_p=1|A_p^{'}=1, Y=1)\neq p(\tilde{A}_p=1|A_p^{'}=0, Y=1)$

(2) This assumption is also the relaxation of the  assumption(2) of Theorem~\ref{theo1}, i.e., $p(E,A_p^{'}|\tilde{A}_p, Y=1) = p(E|\tilde{A}_p, Y=1)p(A_p^{'}|\tilde{A}_p, Y=1)$; 

 We then prove Theorem~\ref{theo2} as follows: since  the assumption (2) $p(E,A_p^{'}|\tilde{A}_p, Y=1) = p(E|\tilde{A}_p, Y=1)p(A_p^{'}|\tilde{A}_p, Y=1)$  holds, we have $(E\bot A_p^{'}) | (\tilde{A}_p, Y=1)$, then we can have $p(E|A_p^{'},\tilde{A}_p, Y=1)=p(E|\tilde{A}_p, Y=1)$. 
 
 From Theorem~\ref{theo0}, we know when the our  $\mathcal{L}$ of the minimax game of Equation~\ref{eqn:ll} reaches the global optimum, we have $p(E|A_p^{'}=1)=p(E|A_p^{'}=0)$, i.e., $E\bot A_p^{'}$, we can naturally infer that $(E\bot A_p^{'})|Y$. Thus, $p(E|A_p^{'}=1, Y=1)=p(E|A_p^{'}=0, Y=1)$. which is equivalent with  $\sum_{\tilde{A}_p}p(E, \tilde{A}_p|A_p^{'}=1, Y=1)=\sum_{\tilde{A}_p}p(E, \tilde{A}_p|A_p^{'}=0, Y=1)$.
 Therefore,
$
\sum_{\tilde{A}_p}p(E|\tilde{A}_p,A_p^{'}=1, Y=1)p(\tilde{A}_p|A_p^{'}=1, Y=1)=\sum_{\tilde{A}_p}p(E|\tilde{A}_p,A_p^{'}=0, Y=1)p(\tilde{A}_p|A_p^{'}=0, Y=1)\nonumber
$
Also, $p(E|A_p^{'},\tilde{A}_p, Y=1)=p(E|\tilde{A}_p, Y=1)$, thus we can obtain, 
\begin{align}
\sum_{\tilde{A}_p}p(E|\tilde{A}_p)p(\tilde{A}_p|A_p^{'}=1, Y=1)=\sum_{\tilde{A}_p}p(E|\tilde{A}_p)p(\tilde{A}_p|A_p^{'}=0, Y=1)
\end{align}
Based on the above equation and assumption (1) $p(A_p^{'}=1|\tilde{A}_p=1, Y=1)\neq p(A_p^{'}=1|\tilde{A}_p=0, Y=1)$, we can get,
\begin{align}
  & \frac{p(E|\tilde{A}_p=1,Y=1)}{p(E|\tilde{A}_p=0, Y=1)}  \nonumber\\
  &=\frac{p(\tilde{A}_p=0|A_p^{'}=1, Y=1)-p(\tilde{A}_p=0|A_p^{'}=0, Y=1)}{p(\tilde{A}_p=1|A_p^{'}=0, Y=1)-p(\tilde{A}_p=1|A_p^{'}=1, Y=1)} \nonumber\\ 
  &=\frac{(1 - p(\tilde{A}_p=1|A_p^{'}=1, Y=1)) - (1 - p(\tilde{A}_p=1|A_p^{'}=0, Y=1))}{p(\tilde{A}_p=1|A_p^{'}=0, Y=1)-p(\tilde{A}_p=1|A_p^{'}=1, Y=1)}
  \nonumber\\
  &=1
\end{align}
Since we already proof that $p(E|A_c=1)=p(E|A_c=0)$ from Theorem~\ref{theo0}, i.e., $E\bot A_c$, we can naturally infer that $(E\bot A_c)|Y$. Thus, $p(E|A_c=1, Y=1)=p(E|A_c=0, Y=1)$. Also, $\hat{Y}=f_{\theta_Y}(E)$, we can get $p(\hat{Y}|\tilde{A}_p=1, Y=1)=p(\hat{Y}|\tilde{A}_p=0, Y=1)$ and $p(\hat{Y}|A_c=1, Y=1)=p(\hat{Y}|A_c=0, Y=1)$, which is the {\sc \textbf{Equal Opportunity}}. Thus, {\sc Equal Opportunity} is achieved when the minimax game of Equation~\ref{eqn:ll} converges to the global optimum. Concluding the proof.
\end{proof}

\end{document}